%% file: root.tex
\documentclass[journal]{IEEEtran}

\usepackage{subfiles}
\usepackage{amsmath}

\usepackage{amsthm}
\usepackage[font=footnotesize,labelfont=normalfont]{caption}
\usepackage{subcaption}
\usepackage{xcolor}
\usepackage{hyperref}
\usepackage{algpseudocode}
\usepackage[linesnumbered,ruled,vlined]{algorithm2e}
\usepackage{amsfonts}
\usepackage{svg}
\usepackage{cite}

\DeclareMathOperator*{\argmin}{arg\,min}
\newtheorem{theorem}{Theorem}
\usepackage{changepage}
\usepackage{siunitx}
\usepackage{tikz}
\usepackage{float}
\usetikzlibrary{fit,calc}

\colorlet{mypink}{red!40}
\colorlet{myyellow}{yellow!70}
\colorlet{myblue}{cyan!60}

\begin{document}

\title{\LARGE \bf
Hierarchical Trajectory (Re)Planning for a Large Scale Swarm
}

\author{Lishuo Pan, Yutong Wang, and Nora Ayanian
\thanks{All authors are with the Department of Computer Science, Brown University, Providence, RI 02912, USA.
        Email: {\tt\footnotesize \{lishuo\_pan, yutong\_wang5, nora\_ayanian\}@brown.edu}}%
}

\markboth{IEEE TRANSACTIONS ON ROBOTICS}%
{}

\maketitle

\begin{abstract}
We consider the trajectory replanning problem for a large-scale swarm in a cluttered environment. Our path planner replans for robots by utilizing a hierarchical approach, dividing the workspace, and computing collision-free paths for robots within each cell in parallel. Distributed trajectory optimization generates a deadlock-free trajectory for efficient execution and maintains the control feasibility even when the optimization fails. Our hierarchical approach combines the benefits of both centralized and decentralized methods, achieving a high task success rate while providing real-time replanning capability.
Compared to decentralized approaches, our approach effectively avoids deadlocks and collisions, significantly increasing the task success rate. We demonstrate the real-time performance of our algorithm with up to 142 robots in simulation, and a representative 24 physical Crazyflie nano-quadrotor experiment.
\end{abstract}

\section{Introduction}
\subfile{sections/introduction.tex}

\section{Related Work}
\subfile{sections/related_work.tex}

\section{Preliminaries}
\subfile{sections/preliminaries.tex}

\section{Problem Formulation}
\subfile{sections/problem_formulation.tex}

\section{Geometric Partitioning}
\subfile{sections/partition.tex}

\section{Multi-commodity Flow with Optimal Detour}
\subfile{sections/MCF.tex}

\section{Cell-based MAPF Planner}
\subfile{sections/planner.tex}
\section{Failure-tolerant Trajectory Optimization}
\subfile{sections/trajectory_optimization.tex}

\section{Results}
\subfile{sections/results_and_discussion.tex}

\section{Conclusion and Future Work}
\subfile{sections/conclusion_and_future_work.tex}

\bibliographystyle{IEEEtran}
\bibliography{IEEEabrv, refs}

\end{document}

%% file: sections/introduction.tex


Large fleets of robots, 
such as those used in warehouse operations~\cite{wawrla2019applications}, disaster response~\cite{van2020drones}, and delivery~\cite{Scott2017DroneDM}, demand 
coordination solutions that adjust in real time to changing goals.  
In this work, we present a real-time lifelong hierarchical method for navigating a large team of robots to 
independent goals in a large, cluttered environment that 
provides a high task success rate. 
By lifelong, we mean robots can enter and exit the space, and can receive another goal at any time, as they would in a warehouse or delivery problem.
Our approach 
partitions the space into disjoint cells, allowing the planner to run in parallel within each cell. 
A high-level planner routes robots through the partition, while a low-level anytime multi-agent pathfinding (MAPF) algorithm and a trajectory optimization algorithm navigate robots to 
local goals within each cell in parallel. 
The real-time property holds as long as there are not too many cells or robots in a workspace; the limits for real-time operation are empirical and problem-specific, however, we demonstrate real-time performance in simulation for $142$ robots with a $12$-cell partition.

We are particularly interested in unmanned aerial vehicles (UAVs) operating in 3D space, such as in city-scale on-demand package delivery, however, our approach applies to robots operating in 2D as well.
We present two approaches for high-level planning depending on the problem's requirements: 1) an egocentric greedy approach that always operates in real-time and 2) a novel high-level planner that routes robots through the partition using a multi-commodity flow (MCF)~\cite{10.5555/137406} based routing algorithm. 
There are tradeoffs between these two approaches.
The egocentric greedy planner operates in real-time regardless of the number of cells; however, it has no mechanism for reducing congestion, thus it can 
result in expensive low-level planning when many robots are present within some cells. 
On the other hand, the MCF-based approach eases cell congestion by regulating the flow of robots into each cell while ensuring bounded-suboptimal inter-cell routing; thus, it can be useful in environments such as urban UAV package delivery, where different types of cells (e.g., residential vs.\ highway) may have different limits on the influx of robots. The influx of a cell is defined as the anticipated total number of robots entering the cell across the current high-level plan. The MCF-based planners can operate in real-time under certain conditions, thus allowing for lifelong replanning while reducing congestion, which leads to faster,  real-time trajectory planning within each cell. 

\begin{figure}[t]
    \centering
    \includegraphics[width=0.48\textwidth]{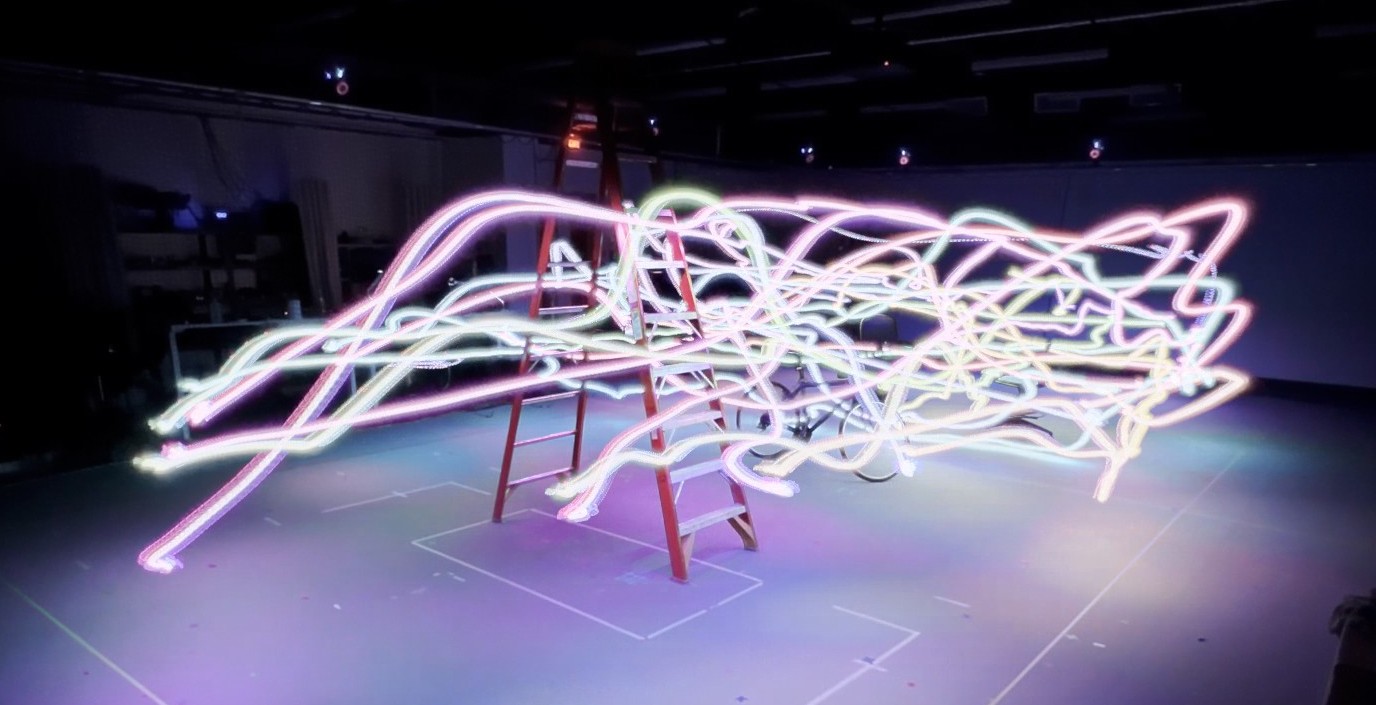}
    \caption{Long exposure of $24$ quadrotors navigating a cluttered environment.}
    \label{fig:demo}
\end{figure}

The MAPF planner computes deadlock-free, collision-free paths while respecting the robots' geometric shape in each cell. Trajectory optimization generates smooth and continuous trajectory for robots to execute while guaranteeing collision avoidance. Rarely, the optimization problem becomes infeasible and a relaxed optimization, without a collision-free guarantee, is solved to maintain the control feasibility. An optional cell-crossing protocol allows robots to transition between cells without stopping in midair. 
Our hierarchical approach allows real-time and deadlock-free execution of multi-robot trajectory planning with a comparable success rate to an offline centralized approach.
The contributions of this work are:
\vspace{-2pt}
\begin{itemize}
    \item  a hierarchical framework for large-scale real-time swarm coordination that significantly reduces planning time compared to the centralized planner and increases the task completion rate compared to decentralized planners; 
    \item  novel multi-commodity flow-based inter-cell routing planners, MCF/OD and one-shot MCF, that reduce congestion by regulating the influx of each cell; and
    \item a failure-tolerant trajectory optimization that generates deadlock-free trajectories and maintains control feasibility when a collision-free guaranteed trajectory planning fails.
    \vspace{-2pt}%
\end{itemize}
We demonstrate the algorithm in simulation with up to $142$ robots and in physical robot experiments with $24$ Crazyflie nano-quadrotors in a cluttered environment, shown in Fig.~\ref{fig:demo}.


%% file: sections/related_work.tex

Centralized approaches to multi-robot trajectory planning~\cite{honig2018trajectory, tajbakhsh2023conflict, guo2021continuous} generate high quality collision- and deadlock-free trajectories.
Search-based MAPF with posterior trajectory optimization~\cite{honig2018trajectory} is complete, but centralized search-based planners like MAPF quickly become intractable with many robots in the system. 
Conflict-based MPC~\cite{tajbakhsh2023conflict} optimizes actions and generates conflict-free trajectories in the prediction horizon; while it reduces computation since the control input is only optimized within a horizon, it still faces computational challenges as the number of robots increases. Additionally, it loses guarantees on task completion due to the limited planning horizon. Continuous-time Gaussian process trajectory generation ~\cite{guo2021continuous} has superior trajectory quality, but they face substantial computational challenges due to the theoretical hardness of the problem in the joint configuration space~\cite{yu2013structure}. This computational hardness prohibits their use for on-demand many-robot replanning. 

On the other hand, decentralized or online algorithms can result in deadlocks, livelocks, collisions, and congestion, thus reducing execution efficiency. 
RLSS~\cite{csenbacslar2023rlss} generates trajectories based on a single-robot planner and a decoupled posterior trajectory optimization without the deadlock-free guarantee. Control barrier functions~\cite{wang2017safety}, or reactive control synthesis methods, are prone to deadlocks. Distributed MPC (DMPC)~\cite{luis2020online} treats collision avoidance as a soft constraint in the cost function, resulting in collisions in the presence of narrow corridors and many robots. 
In cluttered environments, Zhou et al.'s buffered Voronoi cell (BVC)-based algorithm~\cite{zhou2017fast} is prone to deadlocks, and an algorithm using relative safe flight corridor (RSFC)~\cite{park2020online} leads to deadlocks with many robots. 
Recently, the authors'  hierarchical approach~\cite{pan2024hierarchical} attempts to address the computational challenge in many-robot path planning while maintaining a high task success rate similar to centralized approaches, but does so in discrete space. 

In this work, we propose a hierarchical approach to trajectory planning for large-scale swarms in cluttered environments, combining the computation benefits of a decentralized planner, that is, real-time replanning, and task performance benefits of a centralized planner, such as being deadlock-free and having a high task success rate.


%% file: sections/preliminaries.tex

\subsection{Trajectory Parameterization}
\label{sec:preliminaries_bezier}
Quadrotor dynamics with position and yaw inputs have been shown to be \textit{differentially flat}~\cite{mellinger2011minimum}. In the present work, we assume the yaw to be $0$; under this assumption, trajectories are smooth curves $f: \left[ 0, T \right] \rightarrow \mathbb{R}^{3}$ parameterized by time, with duration $T$. To impose smoothness for the entire trajectory, we choose Bézier curves as piecewise splines to easily compute derivatives up to a user-defined degree. A Bézier curve of arbitrary degree $p$ and duration $\tau$ is defined by $p+1$ control points $\boldsymbol{\mathcal{U}} = \left\{\boldsymbol{u}_{0},\cdots,\boldsymbol{u}_{p}\right\}$. We first construct Bernstein polynomials $\boldsymbol{B}_{v}\in \mathbb{R}$ of degree $p$:
\begin{align}
    \boldsymbol{B}_{v} = \binom{p}{v} \left(\frac{t}{\tau}\right)^{v} \left(1-\frac{t}{\tau}\right)^{p-v} \forall t\in \left[0,\tau\right],
\end{align}
where $v=0,1,\cdots,p$.
A $3$-dimensional Bézier curve is defined  $f(t) = \sum_{v=0}^{p}\boldsymbol{B}_{v}\boldsymbol{u}_{v}$ with $\boldsymbol{u}_{v}\in\mathbb{R}^{3}$. The finite set of control points $\boldsymbol{\mathcal{U}}$ uniquely characterizes a smooth Bézier curve and are decision variables in our trajectory optimization (Sec.~\ref{sec:traj_optim}).

\subsection{Multi-agent Pathfinding (MAPF) on Euclidean Graph}
\label{sec:preliminaries_mapf}
Consider an undirected graph $\mathcal{G} = \left(V, E\right)$ embedded in a Euclidean space, where each vertex $v\in V$ corresponds to a position in the free space $\mathcal{F}$ and each edge $(u, v) \in E$ denotes a path in $\mathcal{F}$ connecting vertices $u$ and $v$. For $N$ agents, we additionally require the existence of vertices $v^{i}_{g}$ and $v^{i}_{s}$, corresponding to the goal and start positions, $\mathbf{g}^{i}$ and $\mathbf{s}^{i} \in  \mathbb{R}^{3}$ of robot $r^{i}$ (superscript $i$ represents  robot index), respectfully. At each time step $k$, an agent can either move to a neighbor vertex $\left(u^{i}_{k}, u^{i}_{k+1}\right)\in E$ or stay at its current vertex $u^{i}_{k+1} = u^{i}_{k}$, where $u^{i}_{k} \in V$ is the occupied vertex in the $i$-th agent's path at time step $k$. 
To respect vertex conflict constraints,  no two agents can occupy the same vertex simultaneously, i.e., $\forall k, i \neq j : u^{i}_{k} \neq u^{j}_{k}$. To respect edge conflict constraints, no two agents can traverse the same edge in the opposite direction concurrently, i.e., $\forall k, i \neq j : u^{i}_{k} \neq u^{j}_{k+1}  \vee u^{i}_{k+1} \neq u^{j}_{k}$. 
The objective is to find conflict-free paths 
$\pi^{i} = \left[u^{i}_{0}, \cdots, u^{i}_{K-1}\right]$, where $u^{i}_{0} = v^{i}_{s}$ and $u^{i}_{K-1} = v^{i}_{g}$ for all agents, and minimize cost, e.g., the sum over the time steps required to reach the goals of all agents or the discrete makespan $K$. 
\subsection{MAPF for Embodied Agents and Conflict Annotation}
\label{sec:MAPFC}
Many works address MAPF for embodied agents~\cite{li2019multi, walker2018extended, yakovlev2017any}.
We adopt multi-agent pathfinding with generalized conflicts (MAPF/C), due to its flexibility with different robot shapes. Generalized conflicts, different from typical MAPF conflicts, include the additional conflicts caused by the robots' embodiments~\cite{honig2018trajectory}. 
To account for the downwash effect between aerial robots~\cite{yeo2015empirical}, we denote $\mathcal{R}(\mathbf{r})$ as the convex set of points representing a robot at position $\mathbf{r}$, i.e., a robot collision model. We follow the conflict annotation in~\cite{honig2018trajectory} to annotate the graph with generalized conflicts, defined as: 
\vspace{-0.5em}
\begin{align}\nonumber
\operatorname{conVV}(v)= & \left\{u \in V \mid\right. \\\nonumber
& \left.\mathcal{R}(\operatorname{pos}(u)) \cap \mathcal{R}(\operatorname{pos}(v)) \neq \emptyset\right\} \\\nonumber
\operatorname{conEE}(e)= & \left\{d \in E \mid \mathcal{R}^{*}(d) \cap 
\mathcal{R}^{*}(e) \neq \emptyset\right\}\\\nonumber
\operatorname{conEV}(e)= & \left\{u \in V \mid\right.
\left.\mathcal{R}(\operatorname{pos}(u)) \cap \mathcal{R}^{*}(e) \neq \emptyset\right\},
\end{align}
where $\operatorname{pos}(u) \!\! \in \!\! \mathbb{R}^{3}$ returns the position of vertex $u$. $\mathcal{R}^{*}(e)$ is the $\textit{swept}$ collision model representing the set of points swept by the robot when traversing edge $e$. 


%% file: sections/problem_formulation.tex

Consider a time-varying number of homogeneous non-point robots operating in workspace $\mathcal W$, 
which is partitioned into a union of non-overlapping convex polytopic cells. $\mathcal W$ contains obstacles defined as unions of convex polytopes $\mathcal{O}_{1}, \cdots, \mathcal{O}_{N_{obs}}$. The free space is represented as $\mathcal{F} = \mathcal{W} \backslash (\bigcup_{h} \mathcal{O}_{h})  \ominus \mathcal{R}(\mathbf{0})$, where  $\ominus$ denotes Minkowski difference. Robots must reach specified individual goal positions, which change over time, while avoiding collisions with robots and obstacles and obeying maximum cell influx limits $\boldsymbol{\theta}$ (influx refers to the total number of robots entering a workspace cell across the current high-level plan).
A motivating scenario is a multi-UAV package delivery system where the number of UAVs entering some types of airspace must be limited, and UAVs exit or enter the workspace to charge or redeploy. 

\begin{figure}[t]
    \centering
    \includegraphics[width=0.5\textwidth]{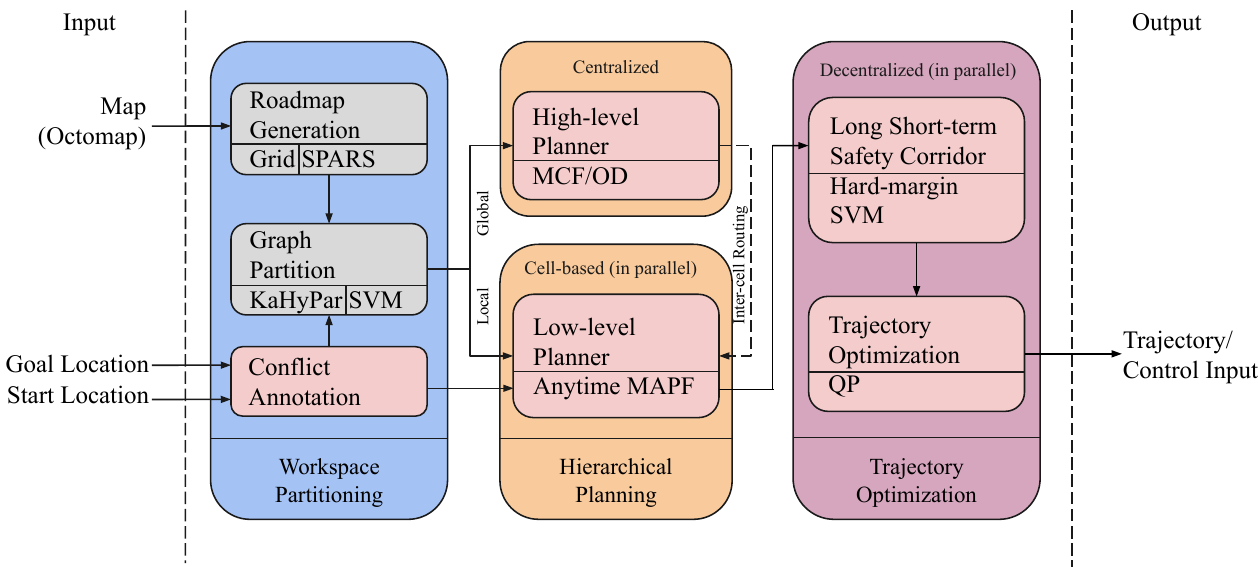}
    \caption{The user inputs a map, start, and goal locations. Our approach generates a geometric partition, distributes robots among  cells (hierarchical planner), and coordinates them within each cell in parallel.\label{fig:outline}}
    \vspace{-2em}
\end{figure}

Our hierarchical trajectory replanning algorithm is divided into three components: workspace partitioning, a hierarchical planner, and optimization-based trajectory planning, as depicted in Fig.~\ref{fig:outline}. The proposed algorithm solves the above lifelong replanning problem in real time and is called repeatedly during execution. 
Partitioning divides the workspace into disjoint convex cells, where the plan can be computed in parallel. Hierarchical planning consists of a centralized high-level planner, i.e., an inter-cell routing algorithm, which obeys cell influx limits while guaranteeing routing quality, and a cell-based anytime planner to compute navigation waypoints within each cell. Trajectory planning generates deadlock-free, collision-free trajectories while respecting the UAVs' dynamic constraints. In rare cases, when solving the optimization is infeasible, our algorithm relaxes the safety constraints to guarantee control feasibility. Despite the relaxation invalidating the safety guarantee, our algorithm maintains a high task success rate, comparable to a centralized approach, without collision even when scaled to a large-scale swarm.     
For each robot $r^{i}$ with initial position $\mathbf{s}^{i} \in \mathbb{R}^{3}$, our algorithm generates an on-demand trajectory $f_{i}(t)$ to the assigned goal position such that there are no collisions between robots or between robots and $\bigcup_{h}\mathcal{O}_h$
and the total number of robots that enter each cell is less than its user-defined influx $\theta_{m}$ (influx limits can vary by cell).

%% file: sections/partition.tex


Here we present our algorithm for partitioning the workspace. Geometric partitioning of a bounded workspace 
consists of three steps: 1) roadmap generation, 2) graph partitioning and spatial linear separation, and 3) local goal generation. 

\subsection{Roadmap Generation}
A roadmap is an undirected graph, introduced in Sec.~\ref{sec:preliminaries_mapf}, 
satisfying three properties: 1)  connectivity-preserving,
i.e., if a path between two points in $\mathcal{F}$ exists, there should be a path in the roadmap as well; 
2) optimality-preserving, i.e., the shortest path between two points in $\mathcal{F}$ can be well approximated by a path in the roadmap; and 3) sparse, i.e., have a small number of vertices and edges. 
In our experiments, we use a 6-connected grid graph, however, it can be generated by other methods, such as SPARS~\cite{dobson2014sparse}. 
\subsection{Graph Partitioning and Spatial Linear Separation}
Our method partitions the workspace into disjoint convex cells and solves a MAPF instance in each cell. Partitioning has two benefits: 1) fewer robots for each subgraph, and 2) decomposed instances can be solved in parallel.
While the user can adopt any  workspace partitioning approach, for example, they may desire to differentiate between residential and commercial areas, 
we propose a method that generates $Q$ convex polytopes. 
First, we use graph partitioning (KaHyPar~\cite{gottesburen2020advanced}) 
to group the roadmap into $Q$ \textit{balanced} (similar number of vertices) subgraphs $\mathcal{G}_{m}\!\!=\!\!\left({V_{m}, E_{m}}\right)$, for $m = 1,\ldots,Q$. 
\textit{Balanced} subgraphs lead to cells of similar volume and can more evenly spread robots within the workspace.

We further enforce each cell, containing a subgraph, to be a convex polytope. Cell convexity prevents robots from penetrating into neighboring cells before exiting their current cells.
We use soft-margin support vector machines (SVM)~\cite{cortes1995support} to compute a hyperplane $\mathcal{H}_{ml} \!\!:=\!\! \left\{ \boldsymbol{w}^{\top}_{c}\mathbf{p} + b_{c} = 0 \right\}$ between vertices of $\mathcal{G}_{m}$ and  $\mathcal{G}_{l}$, where vertices from $\mathcal{G}_{m}$ are on its negative side, and reassign misclassified vertices. Here, the subscript $l$ denotes the index of the subgraphs with vertices connecting to vertices of $\mathcal{G}_{m}$. The resulting set of hyperplanes forms the $m$-th cell, denoted as $P_{m}$. 
\subsection{Local Goal Generation}
For navigating out of a cell, we generate candidate goal states on the faces between adjacent cells. For each cell $P_{m}$, we uniformly sample random local goals on the hyperplane $H_{ml}$ and add them as shared vertices to both $\mathcal{G}_{m}$ and $\mathcal{G}_{l}$. Despite being shared vertices, local goals generated by partition $P_{m}$ have in-edges from $P_{m}$ and out-edges to $P_{l}$ to avoid collision during cell transit. Thus, this part of the graph is directed. 
To enable parallel computation,  there must be no communication between cells. Thus, the cell roadmap $\mathcal{G}_{m}$ must be modified such that the planned paths are collision-free when crossing cells without information exchange between cells. The following properties must be satisfied:
\noindent\textbf{P1}: Robots avoid collision when stationary at vertices (local-goal or non-local-goal vertices) of different cells (generalized vertex-vertex conflict across cells), i.e., $\forall i, j, m \neq  l : v^{i}_{m} \!\!\notin\!\! \operatorname{conVV}(v^{j}_{l})$, where the superscript in $v^{i}_{m}$ refers to the vertex index and the subscript refers to the cell index.


\noindent\textbf{P2}: Robots avoid collision when traversing edges 
between different cells (generalized edge-edge conflict across cells), i.e., $\forall i, k, m \neq l: e^{i}_{m}\!\!:=\!\!(v^{i}_{m}, v^{j}_{m}) \notin \operatorname{conEE}(e^{k}_{l})$, here the superscript in $e^{i}_{m}$ refers to the edge index. 

\noindent\textbf{P3}: Robots avoid collision when one robot is stationary at a vertex while the other robot is traversing an edge of a different cell (generalized edge-vertex conflict across cells), i.e., $\forall i, k, m \neq  l: e^{i}_{m}\!\!:=\!\!(v^{i}_{m}, v^{j}_{m}), v^{k}_{l} \notin \operatorname{conEV}(e^{i}_{m})$.

\begin{figure}[t]
    \begin{subfigure}{0.115\textwidth}
       \centering {\footnotesize(a)}\includegraphics[height=.95\textwidth]{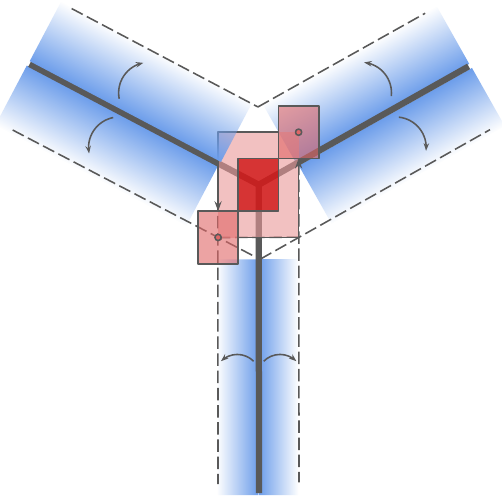}
    \label{fig:partition_generalized_conflict_1}
    \end{subfigure}
    \begin{subfigure}{0.115\textwidth}
        \centering {\footnotesize(b)}\includegraphics[height=.95\textwidth]{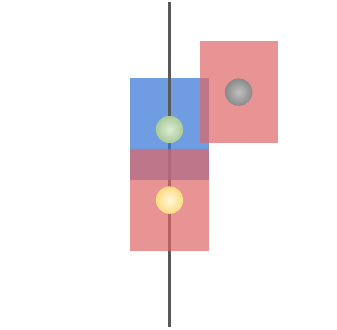}
    \label{fig:partition_generalized_conflict_2}
    \end{subfigure}
    \begin{subfigure}{0.115\textwidth}
        \centering {\footnotesize(c)}\includegraphics[height=.95\textwidth]{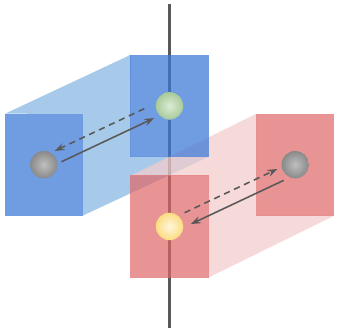}
    \label{fig:partition_generalized_conflict_3}
    \end{subfigure}
    \begin{subfigure}{0.115\textwidth}
        \centering {\footnotesize(d)}\includegraphics[height=.95\textwidth]{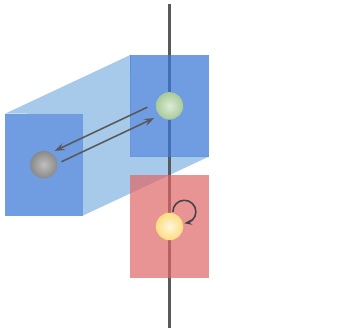}
    \label{fig:partition_generalized_conflict_4}
    \end{subfigure}
    \vspace{-1em}
\caption{(a) inter-robot collision configuration $\mathcal{C}_{col}$ and buffered hyperplanes. (b)-(d) vertex-vertex, edge-edge, and edge-vertex generalized conflicts across cells.\label{fig:partition_generalized_conflict}} 
\vspace{-1.0em}
\end{figure}

\begin{figure}[h]
    \hspace{3em}
    \begin{subfigure}{0.13\textwidth}
       \centering {\footnotesize(a)}\includegraphics[height=.95\textwidth]{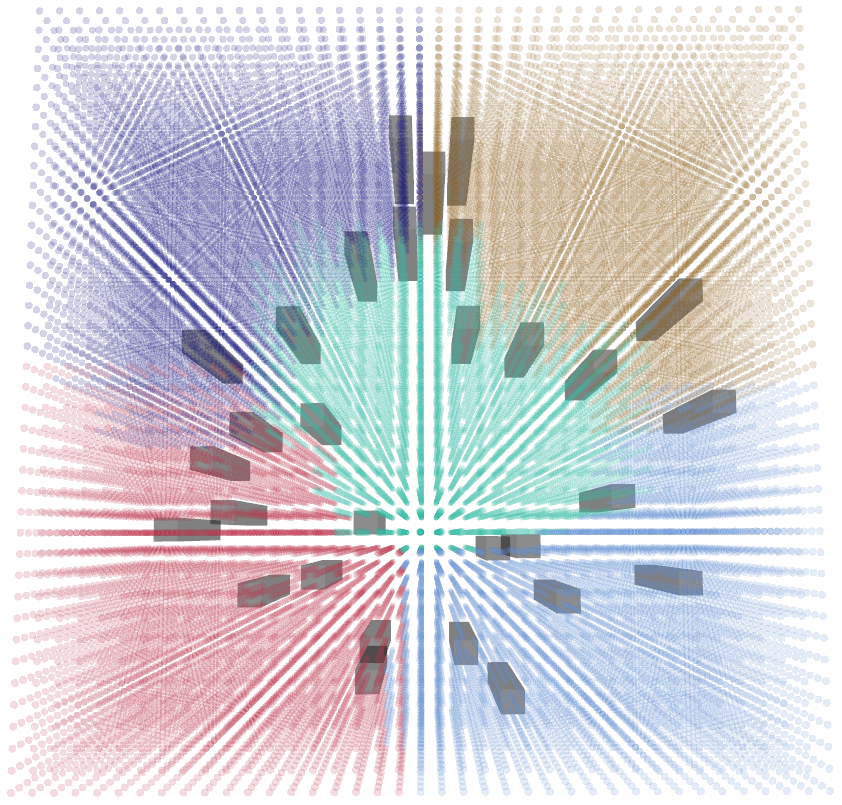}
    \end{subfigure}
    \hspace{2em}
    \begin{subfigure}{0.13\textwidth}
        \centering {\footnotesize(b)}\includegraphics[height=.95\textwidth]{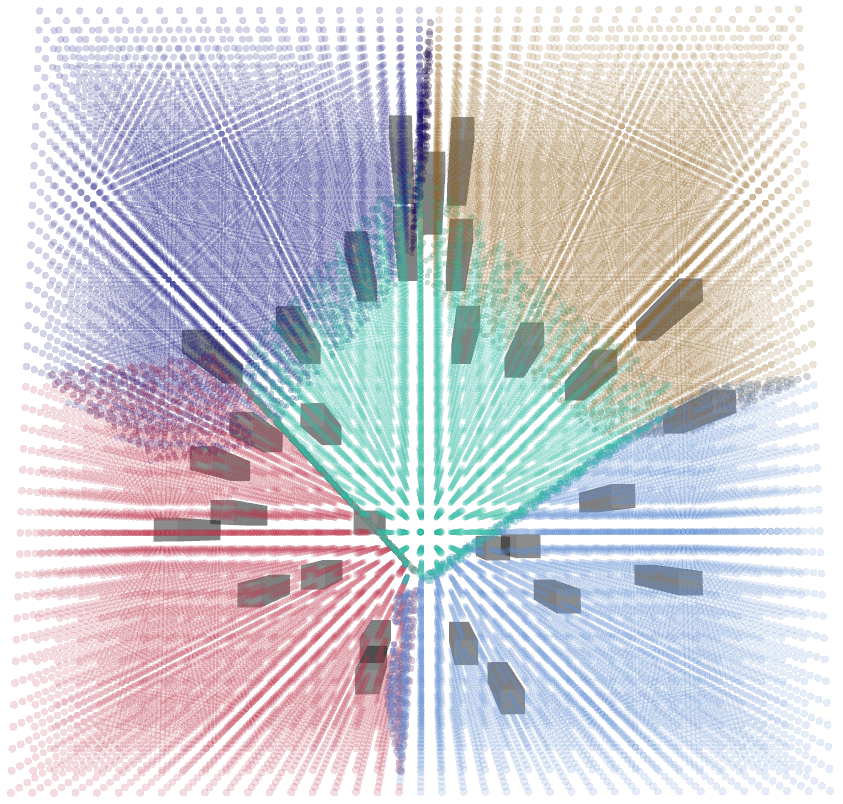}
    \end{subfigure}
\caption{Geometric partitioning with $Q\!\!=\!\!5$, before spatial linear separation (a), and after (b). Each cell is a convex polytope in the workspace.\label{fig:geometric_partition}} 
\vspace{-1.5em}
\end{figure}

We depict violations of P1-P3 in Fig.~\ref{fig:partition_generalized_conflict}b-\ref{fig:partition_generalized_conflict}d. To prevent conflicts between stationary robots at local-goal and non-local-goal vertices across cells, we buffer the separating hyperplane by the inter-robot collision configuration, $\mathcal{C}_{col}$ (c.f. Fig.~\ref{fig:partition_generalized_conflict}a), which is computed as $\mathcal{C}_{col}(\mathbf{p}) = \mathcal{R}_{\mathcal{R}}(\mathbf{p}) \oplus \mathcal{R}_{\mathcal{R}}(\mathbf{0})$, where $\oplus$ is the Minkowski sum. The buffering is achieved by modifying the offset $b_{c}^{'} = b_{c} + \mathrm{max}_{\mathbf{y}\in \mathcal{C}_{col}(\mathbf{0})} \boldsymbol{w}_{c}^{\top} \mathbf{y}$ of the hyperplane.
Given the buffered hyperplanes, all non-local-goal vertices within the buffered region are removed, preventing collisions between stationary robots at local-goal and non-local-goal vertices. The local goals sampled on the hyperplane are confined within the blue region in Fig.~\ref{fig:partition_generalized_conflict}a to avoid vertex-vertex conflicts between stationary robots at local goals of different separating hyperplanes. 
The buffering satisfies P1 between local-goal and non-local-goal vertices across cells and P1, P2, and P3 between non-local-goal vertices across cells. 
To satisfy P1 between local-goal vertices across cells, we uniformly randomly sample points on the hyperplane and reject those that violate P1. 
For P2 and P3 between local-goal and non-local-goal vertices, we add the connection between the sampled local-goal vertex to valid (no collision with the environment) non-local-goal vertices within a radius on both sides of the hyperplane. The local goal is removed if any connected edge violates P2 or P3 (see Fig.~\ref{fig:partition_generalized_conflict}c,~\ref{fig:partition_generalized_conflict}d). Otherwise, the edges are added to the cell roadmaps. Fig.~\ref{fig:geometric_partition}  depicts an exemplar partition. 

%% file: sections/MCF.tex


\label{sec:MCF}
Our hierarchical approach relies on high-level planning, or inter-cell routing, to 1) regulate cell congestion, and 2) preserve the bounded-suboptimality of inter-cell routing solutions. 
Thus, our high-level planner simplifies  MAPF instances within cells and facilitates real-time replanning.
We abstract the partition as a directed graph $\mathcal{G}_{p}\!\!=\!\!\left(V_{p}, E_{p}\right)$, where the vertices (nodes) represent cells and edges connect neighboring cells that share at least one face. Edges are weighted according to Euclidean distance between the cells' centers of mass.
The high-level planner finds an inter-cell routing $\mathcal{T}^{i}\!\!=\!\! \left[ P^{i}_{s}, \cdots, P^{i}_{g}\right]$ for each robot $r^{i}$, where $P^{i}_{s}$ and $P^{i}_{g}$ are its start and goal cell, satisfying: 1) the influx of cell $m$ is under a user-defined value $\theta_{m}$, and 2) $cost(\mathcal{T}^{i})\!\!\leq\!\! w_{\mathrm{mcf}}\cdot cost(\mathcal{T}^{i,*})$. Here, $w_{\mathrm{mcf}} \!\!\geq\!\! 1$ is a scalar representing the suboptimality bound for the routing solutions, and $\mathcal{T}^{i,*}$ is the optimal single-robot routing of robot $r^{i}$. 
The influx of a cell is defined as the anticipated total number of robots entering the cell across the current high-level plan; we define influx formally in the following section.
\subsection{High-level Planning Formulation}
\label{sec:high_level_formulation}
The SOTA partition-based MAPF solvers~\cite{zhang2021hierarchical, leet2022shard} suffer from cell congestion,  which leads to hard instances in certain cells that cause computational bottlenecks. To address this, we formulate the inter-cell routing as a variant of the MCF problem and propose multi-commodity flow with optimal detour (MCF/OD), which optimally distributes robots among cells such that the total number of robots entering any intermediate cell $m$ (not the start or goal cells of the robot teams) across the high-level plan is under a user-defined value $\theta_{m}$, if a solution exists.

Specifically, robots sharing the same start cell $P_{s}$ and goal cell $P_{g}$ are one commodity $c_{sg}$. The commodities set $\mathcal{C}\!\!=\!\!\left\{c_{1}, \cdots, c_{O}\right\}$ includes all commodities given the robot positions. Solving the MCF problem results in optimal flows $\{ y^{*}_{sgml}\}$, that is the number of robots in commodity $c_{sg}\!\!\in\!\! \mathcal{C}$ traversing along edge $e_{ml}\!\in\! E_{p}$.    
In our minimal influx MCF formulation, the optimal flow solutions lead to minimized intermediate cell influx and the most dispersed routing. 
We define the cell influx of $P_{l}$ as the total number of entering robots, that is, $\sum_{c_{sg}\in \mathcal{C}, e_{ml}\in E_{p}} y_{sgml}$.
We formulate the following integer linear program (ILP):
\begin{subequations}
\begin{IEEEeqnarray}{rCl'rCl'rCl}
\argmin_{  \left\{y_{sgml}\right\}} \quad \alpha \!&~&\cdot \! \sum_{c_{sg}}\mathcal{L}_{sg} + \beta \! \cdot \! \mathcal{L}_{in}   \\
\text{s.t.}\sum_{e^{ml},e^{ln}\in E_{p}} \!\!\!\!&&y_{sgml} - y_{sgln} = 
    \begin{cases} 
    |c_{sg}|,  &l = g\\
    -|c_{sg}|,  \!\!\!\!&l=s\\
    0,  &o.w.
    \end{cases}, \forall c_{sg} \in C \label{constr:mcf}\\
\mathcal{L}_{sg} &&\geq y_{sgml},~\forall e_{ml}\in E_{p}, \forall c_{sg} \in \mathcal{C} \label{constr:L_sg}\\
\mathcal{L}_{in} &&\geq \mathbf{I}^{\top}_{v^{i}}\mathbf{y},~\forall v^{i} \in V_{p} \label{constr:L_in}\\
y_{sgml} &&\in \left[0, |c_{sg}|\right],~\forall e_{ml}\in SP_{sg}, \forall c_{sg} \in \mathcal{C}\label{constr:SP_constr1}\\
y_{sgml} &&= 0,~\forall e_{ml}\notin SP_{sg}, \forall c_{sg} \in \mathcal{C}.\label{constr:SP_constr2}
\end{IEEEeqnarray}
\end{subequations}
Here, $\mathcal{L}_{sg}$ represents the maximum flow for commodity $c_{sg}\in \mathcal{C}$, defined by (\ref{constr:L_sg}). $\mathcal{L}_{in}$ represents the maximum influx among all the cells, defined by (\ref{constr:L_in}), where $\mathbf{I}_{v^{i}} = \left[ I_{y_{1}}, \cdots, I_{y_{F}}\right]$, where $y_{f} \in \left\{y_{sgml}\right\}, I_{y_{f}}\in \left\{0, 1\right\}$ is an indicator function that returns $1$ when $y_{f}$ has positive flow to an intermediate vertex $v^{i}\in V_{p}$. $\mathbf{y}$ is the vector of all flows. By minimizing both, the objective function penalizes the maximum congestion among all cells and disperses the flows related to one commodity. We weight $\mathcal{L}_{sg}$ and $\mathcal{L}_{in}$ with coefficients $\alpha$ and $\beta$, respectively, and set $\beta \gg \alpha$ to prioritize minimizing $\mathcal{L}_{in}$. (\ref{constr:mcf}) is the set of constraints for MCF formulation. The set of constraints (\ref{constr:SP_constr1}, \ref{constr:SP_constr2}) enforces the flows $y_{sgml}$ onto the shortest paths between the start and goal cells $SP_{sg}$ guaranteeing the solution optimality.


Despite the minimized influx to the intermediate cells, the  minimal influx MCF formulation leads to congestion in certain cells due to tight constraints on the shortest paths (single-robot shortest routing $\mathcal{T}^{i}$ may intersect at certain cells). To detour the robots optimally, ensuring that the number of robots that enter any intermediate cell $m$ remains below its influx limit $\theta_{m}$ and no unnecessary detour is introduced, we present a complete and optimal solver, MCF/OD, in Algo.~\ref{alg:MCF_optimal_detouring}. It maintains a conflict tree and resolves congestion iteratively. The function $congestionDetection(\mathcal{G}_{p}, P.solution, \boldsymbol{\theta})$ computes the influx for each cell in $\mathcal{G}_{p}$, given the flow solution $P.solution$ and returns the set of cells with their influx larger than the corresponding limit. The function $getAllConflict(P, CS)$ returns the set of all commodities passing congested cells. 
\setlength{\textfloatsep}{0pt}
\begin{algorithm}[t]
\scriptsize
\caption{MCF/OD}
\label{alg:MCF_optimal_detouring}

\KwIn{a multi-commodity flow instance.}
Root.conflict\_counts$[o]$ = $1, \text{for } o = 1,\cdots, O$\\
Root.shortest\_paths = solve shortest path for each commodity in the $\mathcal{G}_{p}$\\ 
Root.cost = sum of the largest cost among all paths in each commodity\\
Root.solution = solve MCF with root shortest path constraints\\ 
insert Root to OPEN\\
Visited[Root.conflict\_counts] = True\\
\While{OPEN not empty} {
    $P \leftarrow$ node from OPEN with the lowest cost\\
    $CS \leftarrow congestionDetection(\mathcal{G}_{p}, P.solution, \boldsymbol{\theta})$\\
    \If{CS is empty} {
        \Return{P.solution}
    }
    $\mathcal{C}_\mathrm{conf} \leftarrow $ $getAllConflict\left(P, CS\right)$.\\
    \For{commodity $c_{o}\in \mathcal{C}_\mathrm{conf}$} {
        $A\leftarrow $ new node.\\
        $A$.conflict\_counts $\leftarrow P$.conflict\_counts + $\mathbf{I}_{o}$\\
        \If{Visited[A.conflict\_counts]} {
            Continue
        }
        $A$.shortest\_paths $\leftarrow P$.shortest\_paths\\
        $p_{k} \leftarrow generateKShortestPath(c_{o}, \mathcal{G}_{p},$ $A$.conflict\_counts$[o])$\\
        Update $A$.shortest\_paths[o] with $p_{k}$ \\
        \If{$p_{k} \leq w_{\mathrm{mcf}} \cdot p_{\mathrm{min}}$} {
            Update $A$.cost\\
            Update $A$.solution by solving MCF with updated shortest paths\\ 
            \If {$A$.cost $\leq \infty$} {
            Insert $A$ to OPEN
        }
        }
    }
}
\end{algorithm}
\begin{theorem}
MCF/OD is complete on a locally finite graph. 
\end{theorem}
\begin{proof}
The cost of a conflict tree node equals the sum of the costs of the longest routing (without cycles) in all commodities. For each expansion, $k$-th shortest paths will be added to the commodity,
which means the cost of the conflict tree is monotonically non-decreasing. For each pair of costs $X < Y$, the search will expand all nodes with cost $X$ before it expands the node with cost $Y$. As the graph is locally finite, there are a finite number of routing with the same cost for each commodity. Thus, expanding nodes with cost $X$ requires a finite number of iterations.

To include an arbitrary combination of $\hat{Z}$ unique edges 
of all commodities, the minimal cost of the conflict tree node is $Z$. $Z$ is finite, as the worst-case scenario is to include all the cell routing within the suboptimality bound. 
Since we are considering a graph with well-defined edge weights and a finite number of commodities, the worst cost is finite. 
For a finite cost $Z$, because the conflict tree node cost is monotonically non-decreasing and only a finite number of nodes with the same cost exists, we can find arbitrary combinations of $\hat{Z}$ unique edges in finite expansions. Thus, if a solution exists by including a combination of $\hat{Z}$ unique edges in the MCF, the algorithm can find it within finite expansions.
If all the unique edges have been added to the MCF solver and the optimization cannot find the solution that satisfies the user-defined influx limit, the problem is identified as unsolvable. 
\end{proof}
\begin{theorem}
MCF/OD is optimal. If a solution is found, it will have the lowest possible cost, i.e., the sum of the costs of the longest routing in all commodities will be minimized if a solution is found.
\end{theorem}
\begin{proof}
MCF/OD is a best-first search. In each expansion, the $k$-th shortest path for the selected commodity is inserted. Thus, the cost of a descendant node is monotonically non-decreasing. Therefore, if a solution is found, it is the optimal solution w.r.t. the cost. 
\end{proof}
MCF/OD can find the optimal detouring solution. However, the complexity is high due to solving an ILP in each expansion. 
To tackle many commodities in a large partition, we propose another efficient detour algorithm, one-shot MCF, which solves MCF once. One-shot MCF augments the shortest paths in \eqref{constr:SP_constr1}, \eqref{constr:SP_constr2} to include all the $w_{\mathrm{mcf}}$ bounded-suboptimal paths for each commodity. We employ the $k$-th shortest path routing algorithm to find all the candidate paths. The proposed One-shot MCF is complete as it includes all bounded-suboptimal paths for each commodity. While it does not optimize for the routing length for all commodities, it optimizes for minimum influx.
Intuitively, MCF/OD adds bounded-suboptimal paths iteratively to relax the constraints and terminates once all cell influx limits are satisfied. On the other hand, One-shot MCF adds all bounded-suboptimal paths at once and optimizes for the minimum influx, so it could result in unnecessary detour.

In each high-level planning iteration, we run both MCF/OD and One-shot MCF in parallel. If MCF/OD times out, we use the solution generated from One-shot MCF. The high-level replanning happens every $\delta_{h}$ time interval.





%% file: sections/planner.tex


Within each cell, the low-level planner, or the cell planner,
computes deadlock-free, collision-free paths that navigate robots to their local goals in an anytime fashion.
The cell planner can be divided into three steps: 1) assigning local goals 
, 2) generating deadlock-free, collision-free discrete paths with anytime MAPF/C, and 3)  optional cell-crossing protocol for non-stop transiting between cells.

\subsection{Local Goal Assignment}
As the first step, local goal assignment aims to route robots to the closest local goals while spreading out robots optimally 
by solving the following ILP:
\begin{subequations}
\begin{IEEEeqnarray}{rCl'rCl'rCl}
\argmin_{A} \quad \sum_{ij} A_{ij} && \cdot D_{ij} + \alpha \sum_{j} u_{j} + \beta U  \\
\text{s.t.} \quad \sum_{j} A_{ij} && = 1, ~\forall i \\ 
U \geq u_{j} && \geq \sum_{i}A_{ij} - 1, ~\forall j.
\end{IEEEeqnarray}
\end{subequations}
$A_{ij} \!\!\in\!\! \{0,1\}$ indicates if robot $r^{i}$ is assigned to the $j$-th local goal, denoted as $lg^{j}$. $D_{ij}$ is the Euclidean distance between $r^{i}$ and $lg^{j}$. 
Auxiliary variables $u_{j}$ in the objective function minimize the number of robots queueing at local goal $lg^{j}$, prioritizing  filling less congested local goals first. 
Auxiliary variable $U$ in the objective function minimizes the maximum number of robots waiting in queue among all the local goals. 
This leads to evenly routing robots to different local goals to reduce congestion. A local goal is occupied if assigned with at least one robot. Thus, the number of robots waiting in queue for a local goal $lg^{j}$ is $\left(\sum_{i} A_{ij} -1\right)$. 
\subsection{Anytime MAPF/C}
We adopt the SOTA anytime MAPF method, namely LNS~\cite{li2021anytime}, which iteratively improves the solution quality until a solution is needed, to facilitate real-time replanning.
We use ECBS~\cite{barer2014suboptimal} for both the initial and iterative planner as it provides bounded-suboptimal and complete solutions. In the initial planning, we use ECBS with a large suboptimal bound to obtain solutions fast.  In the iterative phase, we narrow the suboptimal bound for an improved solution. 
We replan only for a subset of robots for rapid computation. 
We extend ECBS using MAPF/C to account for the robot embodiment. 

The low-level replanning happens repeatedly with a $\delta_{l}$ time interval.

\subsection{Cell-crossing Protocol}
\begin{figure}[t]
    \centering
    \includegraphics[width=0.2\textwidth]{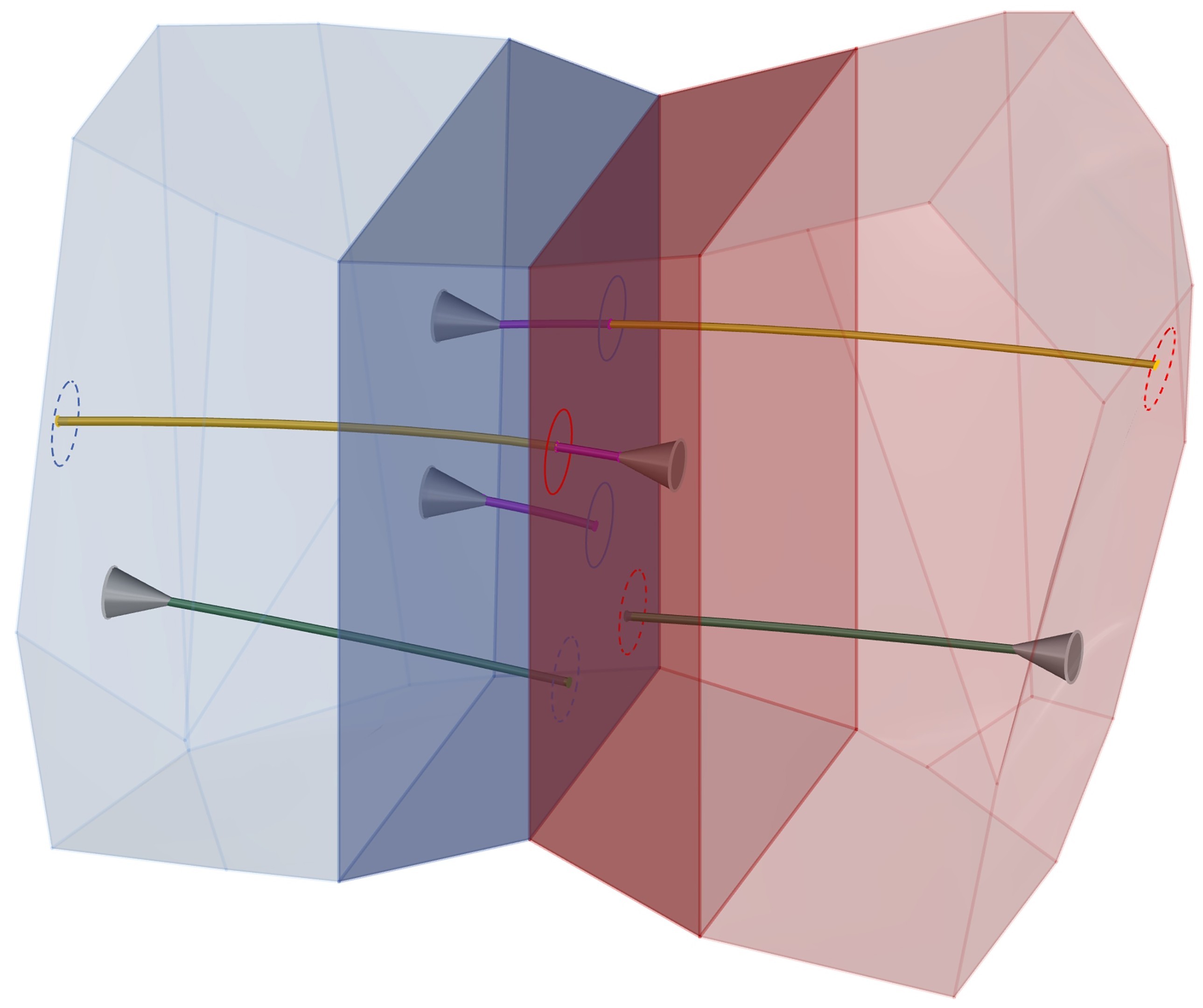}
    \caption{Robots (cones) fix their paths within the buffer zone and start replanning after crossing the hyperplane. 
    }
    \label{fig:between_partition_planning}
\end{figure}

We propose an optional cell-crossing protocol that prevents a robot from idling at its local goal if the path in its next cell is not yet computed. 
We buffer the hyperplane $H_{ml}$ by a distance $d_{e}$ towards cell $P_{m}$. All robots, currently in $P_{m}$ and exiting to $P_{l}$, compute paths for $P_{l}$ within the buffer. 
Buffering is achieved by changing the hyperplane offset to $b^{'}_{c} = b_{c} + \Vert \boldsymbol{w}_{c} \Vert \cdot d_{e}$.
By enforcing the buffer distance $d_{e} \geq \delta_{l} \cdot V_{max}$, where $V_{max}$ is the maximum robot speed, 
the robot is guaranteed to have a plan in its next cell computed at least once 
before leaving its current cell.
A robot entering the buffer zone will then have a plan to exit its current cell and transition through its next cell. The robot, in the buffer zone, fixes its plan of the current cell to lock the local goal and expected arrival time to its next cell. Thus, when computing the plan for the next cell, the robot's expected start time and position will be pre-determined and independent of cell planning order.
The robot computes a plan for its next cell, then concatenates the fixed plan of its current cell and the plan in its next cell to form a complete transition plan. 
Fig.~\ref{fig:between_partition_planning} depicts our cell-crossing protocol.


%% file: sections/trajectory_optimization.tex

\label{sec:traj_optim}
A discrete path has two drawbacks: 1) the robot has to stop at each waypoint, leading to inefficient execution and sharp acceleration when stopping and starting at the waypoint, and 2) the resulting path is resolution-optimal; its optimality is limited on a sparse roadmap and a sparse roadmap is desired in the discrete planning phase for real-time performance. Trajectory optimization addresses these problems. We propose an optimization-based trajectory generation method for robots to navigate the workspace efficiently.

While the proposed method can be applied to other robot platforms, in the present work, we focus on trajectory generation for quadrotors. Quadrotor dynamics with position ($\mathbf{r}$) and yaw ($\psi$) inputs are \textit{differentially flat}, which means the full $12$-DoF state can be expressed with $\left[\mathbf{r}, \psi\right]$ and their derivatives~\cite{mellinger2011minimum}. In this work, we assume the quadrotor yaw $\psi$ is always $0$ during execution. For a discrete path $\pi^{i}$, we assign a time $t_{k} = k\Delta t$ to each discrete time step with a user-defined step duration $\Delta t$. 
The objective is to convert the time-stamped path to a trajectory $f^{i}(t):\left[0, T\right]\rightarrow \mathbb{R}^{3}$ for each robot in the replanning cycle. 
The generated trajectory guarantees deadlock-free execution, avoids collisions with other robots and obstacles, and obeys the quadrotors' dynamic constraints. In rare cases, when the optimization is infeasible due to a sudden change of safety corridors, we relax the safety constraints to maintain control feasibility between replans.


\subsection{Safety Corridor Constraints}
We require the robot to avoid collision with other robots and the environment and assume perfect sensing of other robots and obstacles.  A sufficient condition to achieve collision avoidance under a replanning scheme is to consider safety radii between robots and between robots and obstacles, namely, $d_{r}$ and $d_{e}$. Only other robots and obstacles within the safety radii are considered for potential collision. By considering only the neighbors and obstacles within the safety radii, we significantly reduce the number of constraints in the optimization problem. Let $d_{r} = 2\cdot (|V_{max}| \delta_{l}+r_{\mathcal{R}})$ and $d_{e} = |V_{max}| \delta_{l}+r_{\mathcal{R}}$, where $r_{\mathcal{R}}$ is the radius of the minimum enclosing sphere of the robot collision shape $\mathcal{R}(\mathbf{0})$. We denote the other robots within the safety radius $d_{r}$ by the set $\mathcal{N}^{i}$, and obstacles within the safety radius $d_{e}$ by the set $\mathcal{N}^{i}_{obs}$. 
To impose collision avoidance constraints, we plan the trajectory of robot $r^{i}$, i.e., piecewise splines, within a sequence of safety corridors, represented by convex polytopes $\mathcal{P}^{i}_{k}$, for $k\in \left\{0,\cdots, K-1 \right\}$. 
For $k$-th segment of discrete path $\pi^{i}$, i.e., the path between $k$-th and $(k+1)$-th waypoints, we generate a spline curve within its safety corridor $\mathcal{P}^{i}_{k}$.

A safety corridor $\mathcal{P}^{i}_{k}$ is the intersection of separating hyperplanes: 1) half-spaces separate robot $r^{i}$ from $r^{j}$ during their $k$-th path segment, for $r^{j} \in \mathcal{N}^{i}$ and $j\neq i$; 2) half-spaces separate robot $r^{i}$ from $\mathcal{O}_{h}$, for $\mathcal{O}_{h} \in \mathcal{N}^{i}_{obs}$. We denote them as robot-robot separating hyperplanes $\mathcal{H}_{r}$ and robot-environment separating hyperplanes $\mathcal{H}_{e}$, respectively.
To guarantee safety while reducing  computation, we distinguish between  long and short-horizon safety corridors. For  path segments within the replanning horizon $\delta_{l}$, we consider more constrained short-horizon safety corridors, which consist of both robot-robot avoidance and robot-environment avoidance. For  path segments exceeding the replanning horizon, we plan a long-horizon trajectory to the goal, constrained with only obstacle avoidance. The long-horizon trajectory, while ignoring robot-robot avoidance, provides a deadlock-free trajectory to the goal within the cell and thus prevents the trajectory from being trapped by the environment. 

\begin{figure}[h]
    \centering
    \subfloat[Robot-robot Separation]{\includegraphics[width=0.22\textwidth]{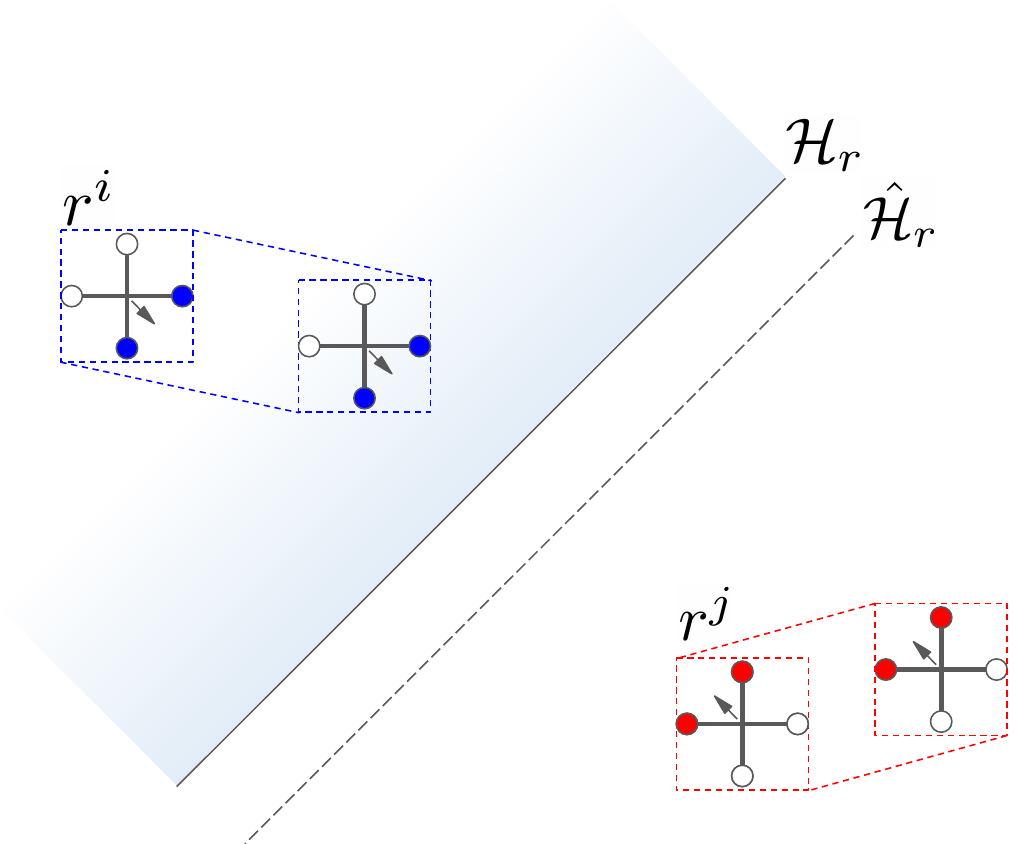}}
    \hspace{1em}
    \subfloat[Robot-environment Separation]{\includegraphics[width=0.22\textwidth]{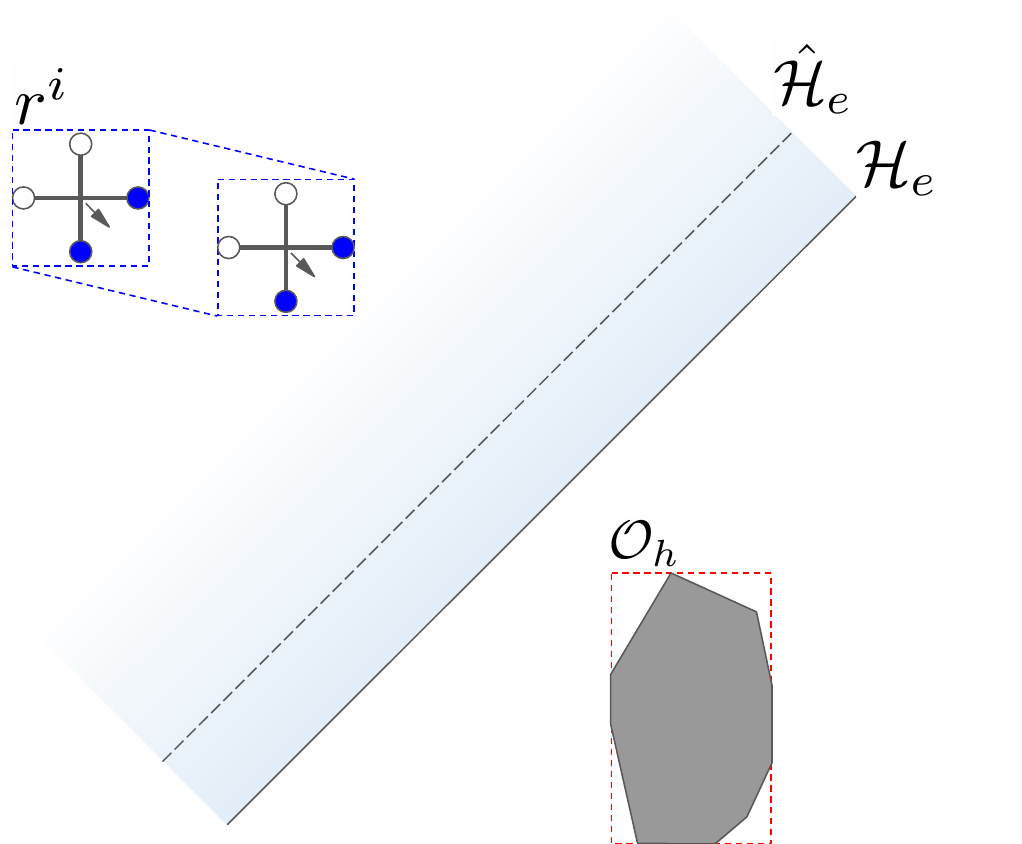}} 
    \caption{Blue robot, red robot, and grey polytope are the ego robot, neighboring robot, and static obstacle, respectively. The dashed bounding boxes are their collision shapes. The blue shaded areas are the corresponding safety half-space.}
    \label{fig:traj_optim_separating_hyperplane_illustration}
\end{figure}

To define separating half-spaces, we introduce the following separating function. The swept space $\mathcal{R}^{*}(e)$ and obstacles $\mathcal{O}$ can be represented as convex sets of points. We use a function $L(\mathcal{A}, \mathcal{B})$ to compute a half-space separating the convex sets of points $\mathcal{A}$ and $\mathcal{B}$. $L(\mathcal{A}, \mathcal{B}) = L(\mathcal{B}, \mathcal{A})$ guarantees the consistency of the generated separating half-space. We define the function $L$ as the hard-margin SVM. For simplicity, we use axis-aligned bounding boxes for the robot's collision shape in the following discussion. We represent robot $r^{i}$ traversing a line segment as a convex hull $\mathcal{A}$. We obtain the convex hull $\mathcal{B}$ for its neighbor $r^{j}$. To avoid collision with a neighbor $r^{j}$, we first compute the half-space $\hat{\mathcal{H}}_{r}\!\! :=\!\! \left\{\boldsymbol{w}^{\top}_{r}\mathbf{p} + b_{r} \leq 0\right\}$ according to function $L$. By buffering its offset $b_{r}^{'} = b_{r} + \mathrm{max}_{\boldsymbol{y}\in \mathcal{R}(\mathbf{0})} \boldsymbol{w}_{r}^{\top} \boldsymbol{y}$, we obtain the robot-robot separating hyperplane $\mathcal{H}_{r}$, as illustrated in Fig.~\ref{fig:traj_optim_separating_hyperplane_illustration}a. By generating trajectory at the negative side of $\mathcal{H}_{r}$, the robot $r^{i}$ is guaranteed to avoid collision with robot $r^{j}$. For robot-environment avoidance, an obstacle $\mathcal{O}_{h}$ is represented as a convex hull. We compute the separating hyperplane $\hat{\mathcal{H}}_{e}\!\! :=\!\! \left\{\boldsymbol{w}^{\top}_{e}\mathbf{p} + b_{e} \leq 0 \right\}$. It is unnecessary to reserve the safety corridor for obstacles, as they are static. By buffering the offset $b_{e}^{'} = b_{e} - \mathrm{min}_{\boldsymbol{x}\in \mathcal{O}_{h}} \boldsymbol{w}_{e}^{\top} \boldsymbol{x} + \mathrm{max}_{\boldsymbol{y}\in \mathcal{R}(\mathbf{0})} \boldsymbol{w}_{e}^{\top} \boldsymbol{y}$, we obtain the robot-environment separating hyperplane $\mathcal{H}_{e}$, which maximizes the safety corridor for the robot $r^{i}$, as illustrated in Fig.~\ref{fig:traj_optim_separating_hyperplane_illustration}b.

\begin{figure}[t]
    \centering
    \includegraphics[width=0.4\textwidth]{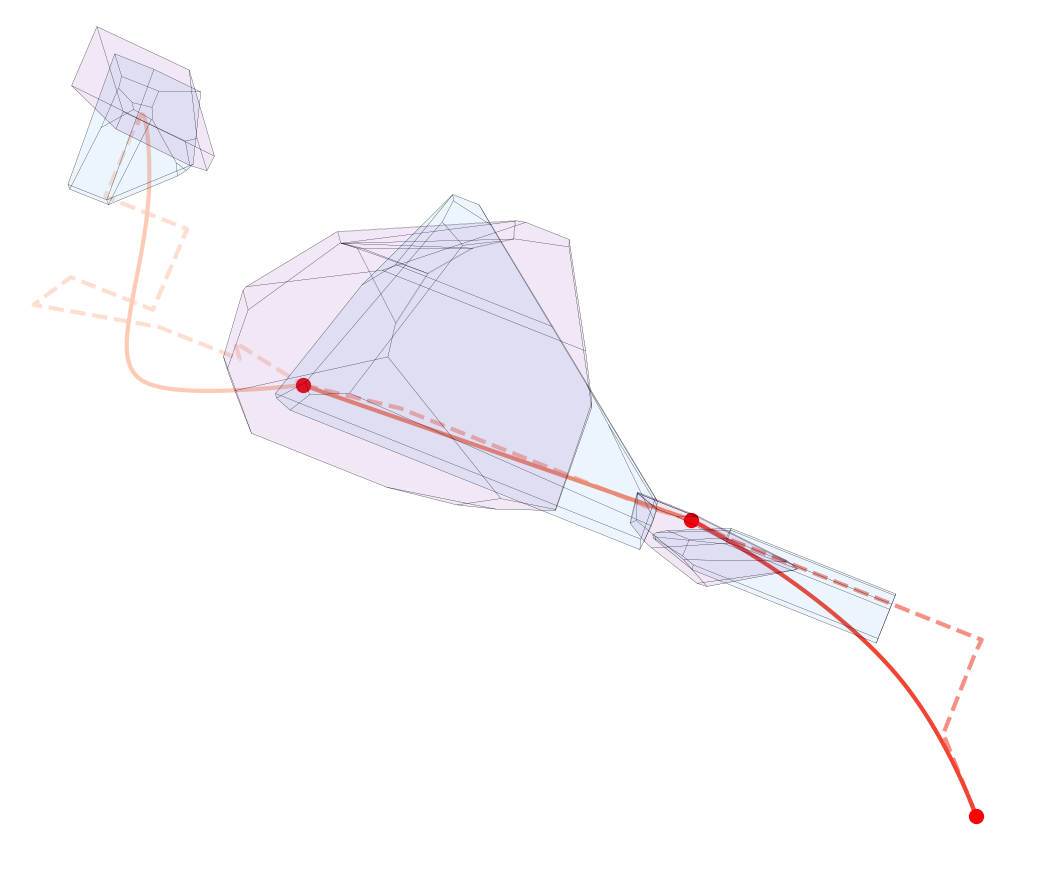}
    \caption{Safety corridors in replanning horizon (long-horizon safety corridors are omitted) at different frames. Dashed lines: discrete paths. Solid curves: Optimized trajectories (colored with temporal information) that reach the cells' local goals(red dots).}
    \label{fig:traj_optim_safety_corridor}
\end{figure}

\subsection{Continuity Constraints}
In the rest of this section, without loss of generality, we only consider the trajectory optimization of one robot. To simplify the discussion, we omit the superscript $i$, which denotes the robot index. The robot's trajectory is represented as a piecewise Bézier curve, where the continuity of control inputs is not guaranteed between curves. To achieve stable control through the entire trajectory, we enforce continuity in higher-order derivatives between consecutive Bézier curves, i.e., $\frac{d^j f_{k}\left(T_{k}\right)}{d t^j}=\frac{d^j f_{k+1}(0)}{d t^j},$ for  $ k \in\{0, \ldots, K-2\}$ and $ j \in\{0, \ldots, C\}$. Here, $f_{k}$ and $T_{k}$ refer to the $k$-th Bézier curve and its duration. $C$ is the degree of derivatives for which we require continuity. 

\subsection{Distributed Trajectory Optimization}
One Bézier curve of arbitrary degree $p$ and duration $\tau$ is defined by $p+1$ control points as described in Sec.~\ref{sec:preliminaries_bezier}. Our algorithm generates piecewise Bézier curves to represent the trajectory, where the $k$-th Bézier curve is constrained within the corresponding safety corridor $\mathcal{P}_{k}$. Our objective is to optimize all the control points $\boldsymbol{\mathcal{U}} = [\boldsymbol{\mathcal{U}}_{0}^{\top}, \cdots , \boldsymbol{\mathcal{U}}_{K-1}^{\top}]^{\top}$, where $\boldsymbol{\mathcal{U}}_{k} = \left[\boldsymbol{u}_{k,0},\cdots,\boldsymbol{u}_{k,p}\right]^{\top}$, such that the generated piecewise Bézier curves satisfy all the above constraints and minimize a cost function. We aim to minimize the weighted sum of integrated squares of derivatives~\cite{richter2016polynomial}, and formulate the following quadratic programming (QP):
\begin{subequations}
\begin{IEEEeqnarray}{rCl'rCl}
\argmin_{\boldsymbol{\mathcal{U}}} 
&~& \sum_{j=1}^{C} \gamma_{j} \int_0^T \left\|\frac{d^{j}}{dt^{j}} f(t;\boldsymbol{\mathcal{U}})\right\|_2^2 \, dt \label{QPcost}\\
\text{s.t.} 
&~& \frac{d^j f(0)}{dt^j} = \frac{d^j \mathbf{r}}{dt^j} \,, \quad \forall j \in\{0, \ldots, C\}, \label{QPconst:init_state}\\
&~& f_{K-1}(T_{K-1}) = \mathbf{r}_{lg} \,, \label{QPconst:final_pos}\\
&~& \frac{d^j f_{K-1}(T_{K-1})}{dt^j} = \mathbf{0} \,, \quad \forall j \in\{1, \ldots, C\}, \label{QPconst:final_high_order}\\
&~& \begin{aligned} \frac{d^j f_{i}\left(T_{i}\right)}{d t^j}=\frac{d^j f_{i+1}(0)}{d t^j}, & \forall i \in\{0, \ldots, K-2\} \\ & \forall j \in\{0, \ldots, C\}\end{aligned} \label{QPconst:continuity}\\
&~& \begin{aligned} A_{i}\boldsymbol{u}_{i,j} + \boldsymbol{b}_{i} \leq 0, & \forall i \in\{0, \ldots, K-1\} \\ & \forall j \in\{0, \ldots, p\}\end{aligned} \label{QPconst:safety_corridor}.
\end{IEEEeqnarray}
\end{subequations}

In the above QP, we aim to minimize the integrated squares of derivatives for the generated trajectory (\ref{QPcost}). At each replanning, we match the trajectory's initial state with the robot's current state (\ref{QPconst:init_state}) and the trajectory's final state with the local goal position and zero high-order derivatives (\ref{QPconst:final_pos},~\ref{QPconst:final_high_order}). To stabilize the control input between consecutive Bézier curves, we enforce  continuity up to derivatives of order $C$ between Bézier curves (\ref{QPconst:continuity}). The $k$-th Bézier curve is generated within the $k$-th polytope $\mathcal{P}_{k}$, satisfied by the linear constraints (\ref{QPconst:safety_corridor}). Any SOTA optimization solver can effciently solve the above QP. A typical trajectory generated by our QP is shown in Fig.~\ref{fig:traj_optim_safety_corridor}.

\subsection{Trajectory Rescaling}
Note that the above QP does not constrain derivative magnitudes, such as maximum acceleration/velocity. Although derivative magnitude constraints can easily be added as linear constraints by bounding the control points, tt has been shown that such constraints can lead to overly conservative derivatives~\cite{mercy2017spline}. Instead, we apply  trajectory rescaling~\cite{honig2018trajectory} to satisfy the dynamic limits. Specifically, we compute the maximum magnitude of all derivatives of the generated trajectory. If any derivative magnitude exceeds the dynamic limit, we scale the piecewise Bézier curves duration by a factor $\gamma$, where $\gamma > 1$. The duration rescaling is repeated iteratively until the dynamic limits are preserved. 

\subsection{Failure-tolerate Trajectory Optimization}

\begin{figure}[h]
    \centering
    \subfloat[Execution time = 32\unit{s}]{\includegraphics[width=0.24\textwidth]{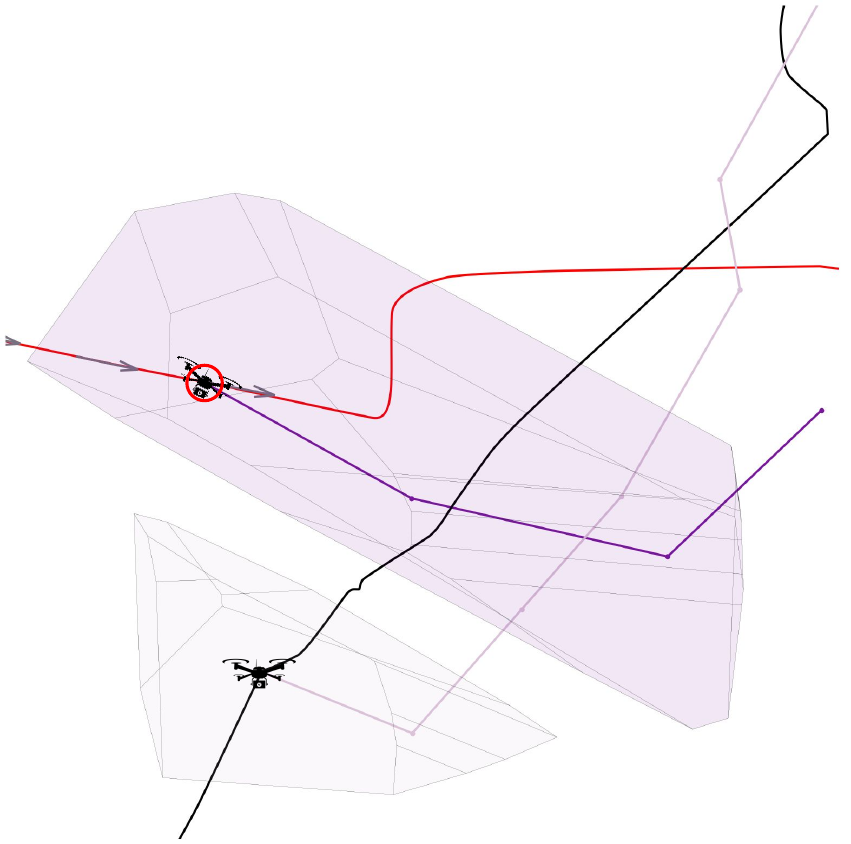}}
    \subfloat[Execution time = 33\unit{s}]{\includegraphics[width=0.24\textwidth]{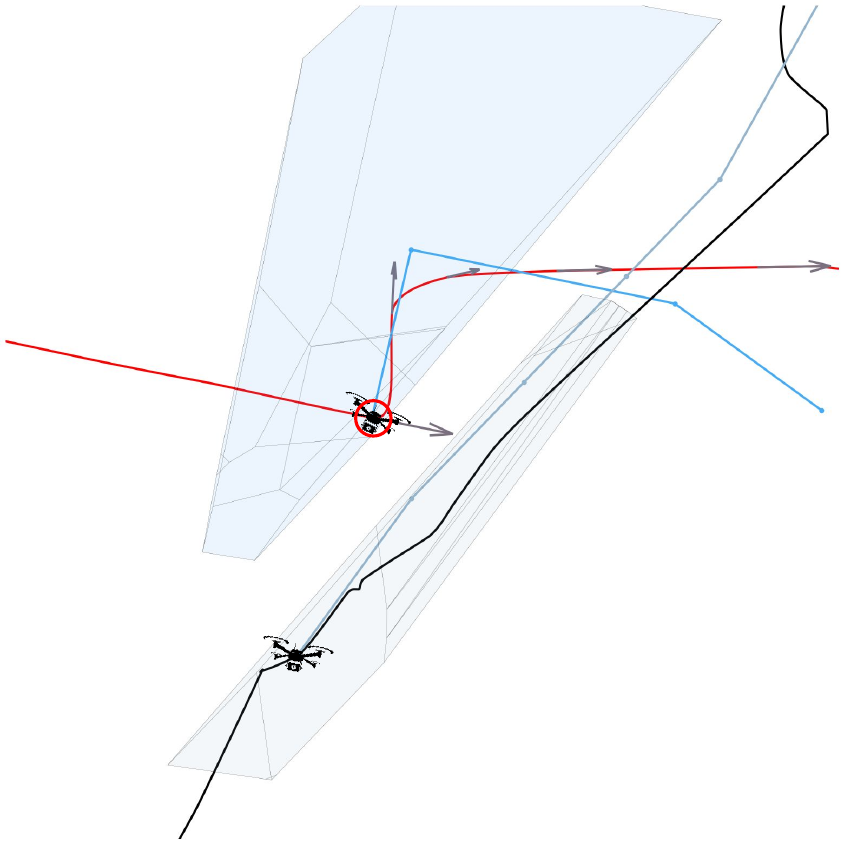}} 
    \caption{The purple solid line is the planned path, and the purple polytopes are the safety corridors of the first path segment for the ego robot and its neighbor at execution time 32$\unit{s}$. Same as the blue solid lines and polytopes at execution time 33$\unit{s}$. The red and black curves are the trajectories of the ego and the other robot. The circled state is where optimization happens. The arrows are the velocities. The sudden change of safety corridors after replanning causes the optimization failure. The relaxed trajectory follows the discrete path.}
    \label{fig:failure_analysis}
\end{figure}

The proposed trajectory optimization is not always successful due to sudden changes in the safety corridor introduced by replanning during execution and the robot's dynamic limits. A robot's path after replanning can deviate from its previous plan even in short horizons, which can lead to a sudden change of its safety corridor, illustrated in Fig.~\ref{fig:failure_analysis}. In that example,  since the robot was gaining velocity in the previous plan, it cannot adapt to dramatic changes in its  flight corridor. This optimization failure happens because we only consider the robot's position and ignore its  derivatives in the discrete planning phase. To maintain feasible control inputs for the robot when optimization fails, we solve a relaxed problem without all the safety corridor constraints in~\eqref{QPconst:safety_corridor} while forcing the robot to pass the discrete waypoint at the end of each Bézier curve, i.e., $f_{k}(T_{k}) = \mathbf{r}_{k}$, where $\mathbf{r}_{k}$ is the position of discrete waypoints. This relaxed trajectory optimizationdoes not provide a safety guarantee, but  we show empirically that collision happens rarely, and our algorithm maintains a high task success rate even with many robots.

%% file: sections/results_and_discussion.tex

\label{sec:results}

We demonstrate the system in experiments on simulated and physical robots. We implement conflict annotation, geometric partitioning, inter-cell routing, discrete planning, and trajectory optimization in C++. We use 
KaHyPar ~\cite{10.1145/3529090} for graph partitioning, Octomap~\cite{hornung2013octomap} for obstacle detection in the physical experiments, Gurobi 10.0 as the ILP solver and CPLEX 12.10 as the QP solver. 

\subsection{Simulation Experiments}
\label{sec:simulation_experiments}

To validate the scalability of our algorithm, we created four instances, namely, ``Circle74", ``Circle142", ``Horizontal48", and ``Swap48", where the number indicates the number of robots. In the ``Circle74" and ``Circle142" instances, robots are initialized in concentric circles with different radii centered at the $x$-$y$ plane's origin and at $1$\unit{m} height with antipodal goals. The column obstacles are randomly generated in the workspace. We scale $x$ and $y$ dimensions by a factor of $\sqrt{N}$ to maintain similar robot density. In the ``Horizontal48" instance, robots are initialized in rectangular grids with $[2.8\unit{m}, 1.4\unit{m}, 1.0\unit{m}]$ spacing in the x, y, and z axes. Robots are required to move horizontally through an environment with column obstacles. In the "Swap48" instance, 
robots are initialized as two 24-robot swarms, starting on the opposite sides of a narrow corridor, arranged in rectangular grids with $[1.2\unit{m}, 2.4\unit{m}, 4\unit{m}]$ spacing in the x, y, and z axes. Robots are required to reach their centrosymmetrically placed goal positions.  In Table~\ref{table:instance}, we specify the workspace, obstacle, and robot start and goal positions.  


\begin{table}[t]
    \centering
    \scalebox{0.75}{
    \begin{tabular}{|c|c|c|c|c|}
        \hline
        Instance & $N$ & Workspace & Obstacles & Start/Goal positions \\
        \hline \hline
        Circle74 & $74$ & $20\unit{m}\!\!\times\!\! 20\unit{m}\!\!\times\!\! 8\unit{m}$ & $20$ columns & Circle radius $10$\unit{m} \\
        \hline
        Circle142 & $142$ & $27.7\unit{m}\!\!\times\!\! 27.7\unit{m}\!\!\times\!\! 8\unit{m}$ & $7$ columns & Circle radii $13.85$, $11.85\unit{m}$ \\
        \hline
        Horizontal48 & $48$ & $27.6\unit{m}\!\!\times\!\! 14.4\unit{m}\!\!\times\!\! 7.2\unit{m}$ & $12$ columns & $2\unit{m} \!\!\times\!\! 8\unit{m} \!\!\times\!\! 3\unit{m}$ grid \\
        \hline
        Swap48 & $48$ & $15.36\unit{m}\!\!\times\!\! 11.52\unit{m}\!\!\times\!\! 10\unit{m}$ & Narrow Corridor & Two $3.6\unit{m} \!\!\times\!\! 4.8\unit{m} \!\!\times\!\! 4\unit{m}$ grids \\
        \hline
    \end{tabular}
    }
    \caption{Configurations of different simulation instances}
    \label{table:instance}
\end{table}

For simplicity, in the high-level planning, we set the influx limits the same for all cells in one instance. We set the influx limit $\theta_{m} = 20$, high-level suboptimal bound $w_{\mathrm{mcf}} = 2$ for ``Circle74", $\theta_{m} = 80$, $w_{\mathrm{mcf}} = 2$ for ``Circle142", $\theta_{m} = 20$, $w_{\mathrm{mcf}} = 2.5$ for ``Horizontal48", and $\theta_{m} = 48$, $w_{\mathrm{mcf}} = 2$ for ``Swap48".  for all cells in high-level planning and a suboptimal bound $w_{\mathrm{mcf}} = 2$ for both MCF/OD and One-shot MCF algorithms. We set the high-level planning time interval $\delta_{\mathrm{h}} = 5$\unit{s} for ``Circle74'',  $\delta_{\mathrm{h}} = 10$\unit{s} for ``Circle142'', $\delta_{\mathrm{h}} = 10$\unit{s} for ``Horizontal48'', and $\delta_{\mathrm{h}} = 10$\unit{s} for ``Swap48''. We set the low-level planning time interval $\delta_{\mathrm{l}} = 1$\unit{s}. The discrete segment duration $\Delta t \!=\! 1/2\unit{s}$. For the LNS planner, we use random neighborhood selection to select the subset of robots to iterate their solutions. We use ECBS/C for both the initial planner and the iterative planner. 
To account for downwash, we use an axis-aligned bounding box from $\left[-0.12\unit{m}, -0.12\unit{m}, -0.2\unit{m} \right]^{\top}$ to $\left[0.12\unit{m}, 0.12\unit{m}, 0.2\unit{m} \right]^{\top}$ to represent the robot collision model $\mathcal{R}(\mathbf{0})$. 
In trajectory optimization, we set the degree of the Bézier curve $p = 7$. We consider continuity up to degree $C = 4$ to minimize the control effort up to snap~\cite{mellinger2011minimum}. Robots have a velocity limit $5$\unit{m/s}, and an acceleration limit $5$\unit{m/s^{2}}, which are chosen arbitrarily. We choose a duration rescaling factor $\gamma = 1.2$. 


\begin{table*}[t]
\centering
\scalebox{0.82}{
\begin{tabular}{c||c||c|c||c|c|c|c||c|c|c||c|c|c}
\multicolumn{1}{c||}{Instance} & \multicolumn{1}{c||}{ Q } & \multicolumn{2}{c||}{ Roadmap } & \multicolumn{4}{c||}{ MCF Routing}  & \multicolumn{3}{c||}{ Discrete Path Planning } & \multicolumn{3}{c}{ Trajectory Optimization } \\
&    & $\left|V\right|$ & $\left|E\right|$  & Method & $\bar{t}_{\mathrm{MCF}}(\unit{s})$ & $\bar{t}^{\mathrm{max}}_{\mathrm{MCF}}(\unit{s})$ & $\bar{N}_{\mathrm{max}}$ & Method & $\bar{t}_{\mathrm{dis}}(\unit{s})$ & $\bar{t}^{\mathrm{max}}_{\mathrm{dis}}(\unit{s})$ & $\bar{t}_{\mathrm{traj}}(\unit{s})$ & $\bar{t}_{\mathrm{traj}}^{\mathrm{max}}(\unit{s})$ & $\bar{T}(\unit{s})$ \\
\hline \hline

Circle74 & 1 & $1110$ & $6728$ & - & - & - & $25.0$ & ECBS(2.0) & $2.71\pm0.05$ & - & $1.17\pm0.01$ & $1.91\pm0.03$ & $\mathbf{35.83}$ \\
\hline
Circle74 & 10 & 1501.8 & 9985.7 & Greedy & & & $19.8$ & ECBS(2.0, 1.8) & $0.02\pm0.00$ & $0.18\pm0.09$ & $0.34\pm0.02$ & $0.68\pm0.03$ & $73$ \\
\hline
Circle74 & 10 & 1504.2 & 10008.1 & MCF & $0.02$ & $0.08$ & \ $\mathbf{14.5}$ & ECBS(2.0, 1.8) & $\mathbf{0.01\pm0.00} $ & $\mathbf{0.13\pm0.06}$ & $\mathbf{0.28\pm0.01}$ & $\mathbf{0.57\pm0.03}$ & $84$ \\
\hline
Circle142 & 1 & $2112$ & $13444$ & - & - & - & $44.0$ & ECBS(4.5) & $12.89\pm0.18$ & - & $2.02\pm0.02$ & $3.34\pm0.14$ & $\mathbf{44.79}$ \\
\hline
Circle142 & 12 & 2925.1 & 19239.0 & MCF & $0.07$ & $0.20$ & \ $\mathbf{23.2}$ & ECBS(4.5, 2.0) & $\mathbf{0.06\pm0.01} $ & $\mathbf{0.32\pm0.06}$ & $\mathbf{0.17\pm0.00}$ & $\mathbf{0.32\pm0.01}$ & $106$ \\
\hline
Horizontal48 & 1   & $2191$ & $13871$ & - & - & - & $38.0$ & ECBS(2.0) & $2.80\pm0.02$ & - & $0.92\pm0.02$ & $1.34\pm0.06$ & $\mathbf{31.10}$ \\
\hline
Horizontal48 & 9 & 2397.3 & 16701.6 & Greedy & & & $22.7$ & ECBS(2.0, 2.0) & $0.14\pm0.04$ & $1.55\pm0.56$ & $0.16\pm0.01$ & $0.29\pm0.02$ & $65$ \\
\hline
Horizontal48 & 9 & 2398.8 & 16715.5 & MCF & $0.01$ & $0.04$ & \ $\mathbf{19.6}$ & ECBS(2.0, 2.0) & $\mathbf{0.04\pm0.02} $ & $\mathbf{0.35\pm0.22}$ & $\mathbf{0.14\pm0.01}$ & $\mathbf{0.27\pm0.02}$ & $66$ \\
\hline 
Swap48 & 1  & $1368$ & $8594$ & - & - & - & $37.0$ & ECBS(3.0) & $3.98\pm0.08$ & - & $0.51\pm0.00$ & $0.87\pm0.01$ & $\mathbf{23.85}$  \\
\hline
Swap48 &7 & 1388.4 & 8810.2 & Greedy & & & $23.1$ & ECBS(3.0, 3.0) & $0.11\pm0.03$ & $1.13\pm0.47$ & $0.15\pm0.01$ & $0.34\pm0.03$ & $37$ \\
\hline
Swap48 & 7 & 1391.4 & 8844.4 & MCF & $0.01$ & $0.03$ & \ $23.1$ & ECBS(3.0, 3.0) & $\mathbf{0.10\pm0.02} $ & $\mathbf{0.69\pm0.16}$ & $0.15\pm0.01$ & $\mathbf{0.33\pm0.02}$ & $38$ \\
\end{tabular}
}
\caption{Experiments on different instances compared to offline planning baseline. The statistics are averaged over $10$ trials.}
\label{table:quantitative}
\vspace{-1.5em}
\end{table*}

\begin{figure*}[h]
    \centering
    \subfloat[Circle74]{\includegraphics[height=0.22\textwidth]{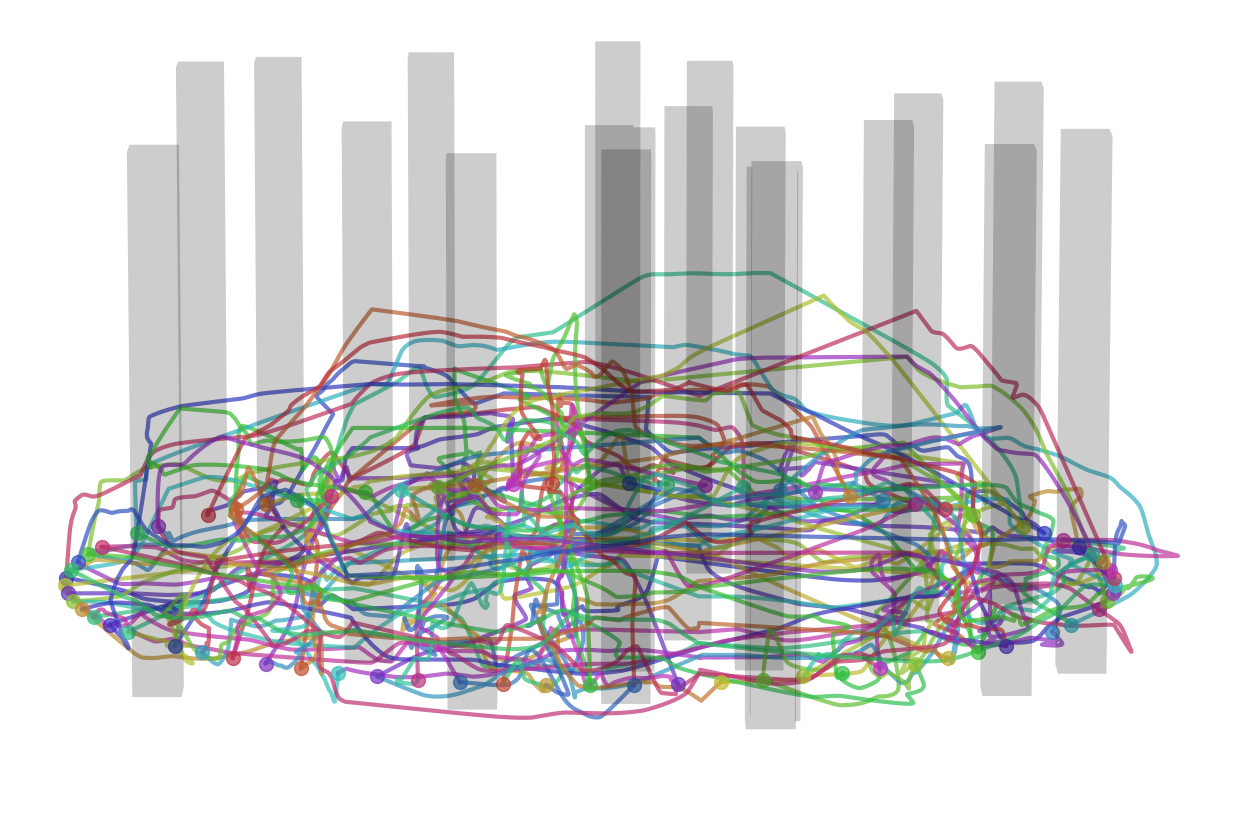}}
    \hspace{1em}
    \subfloat[Circle142]{\includegraphics[height=0.22\textwidth]{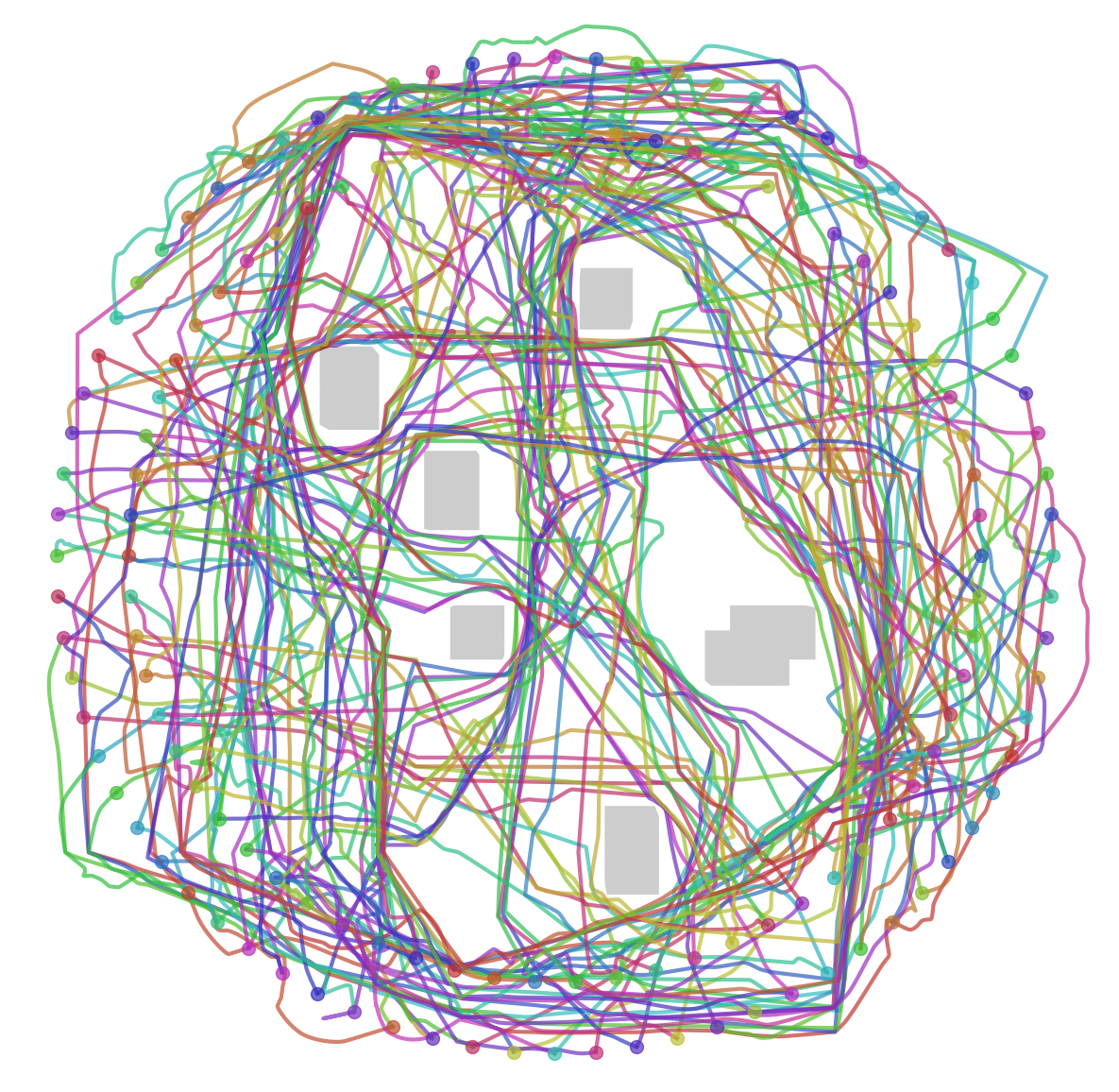}} 
    \hspace{1em}
    \subfloat[Horizontal48]{\includegraphics[height=0.22\textwidth]{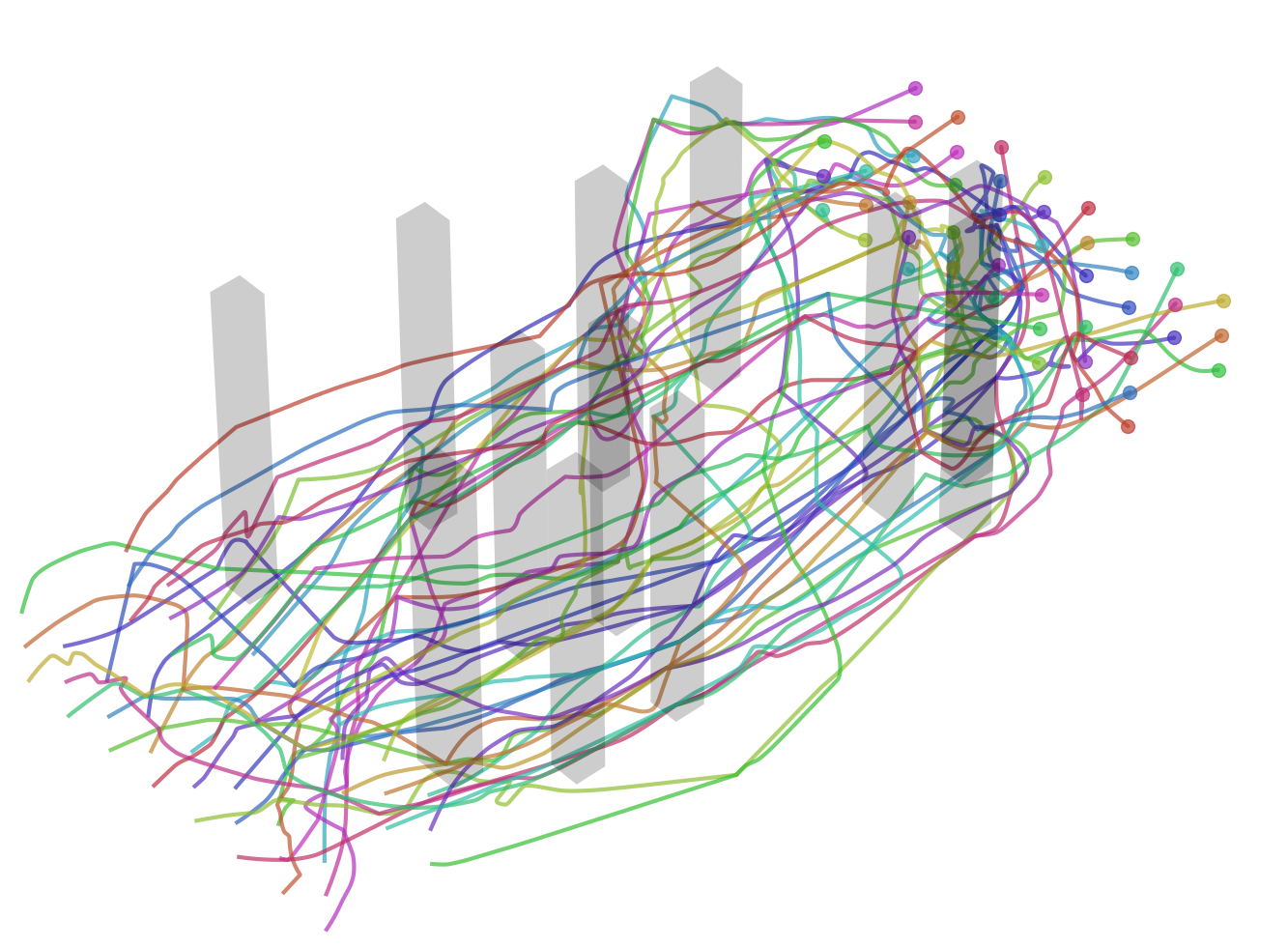}} 
    \caption{Representative trajectories of our algorithm with different instances.}
    \label{fig:qualitative}
\end{figure*}

Fig.~\ref{fig:qualitative} depicts typical solutions of the proposed algorithm. We summarize quantitative results in Table~\ref{table:quantitative}, where we compare to the baseline centralized swarm trajectory planning algorithm~\cite{honig2018trajectory} computed in the same workspace and roadmap. 
Note that, as $\left| V\right|$ and $\left|E\right|$ suggest, the number of vertices and edges increases after partitioning as we add local goals and corresponding edges. $\bar{t}_{\mathrm{MCF}}$, $\bar{t}_{\mathrm{dis}}$ and $\bar{t}_{\mathrm{traj}}$ are the average of computation time of MCF-based inter-cell routing, discrete path planning, and trajectory optimization over trials, and $\bar{t}^{\mathrm{max}}_{\mathrm{MCF}}$, $\bar{t}^{\mathrm{max}}_{\mathrm{dis}}$ and $\bar{t}^{\mathrm{max}}_{\mathrm{traj}}$ are the average of the maximum computation time of the respective components. Despite the simulation being computed in a centralized node, trajectory optimization can be distributed on-board in principle. Here, we compute $\bar{t}_{\mathrm{traj}}$ by averaging over the number of robots. $\bar{N}_{\mathrm{max}}$ is the average maximum number of robots in a cell during execution, and $\bar{T}$ is the average of trajectory makespans.


In comparison with the centralized baseline, the proposed method demonstrates significant computational speed advantages with trade-offs in solution quality. In the MCF routing phase, with a small computational overhead $\bar{t}_{\mathrm{MCF}}$, the proposed method effectively reduces  congestion among cells compared to both greedy and partitionless approaches, by inspecting $\bar{N}_{\mathrm{max}}$. 
As the discrete planning is computed in an anytime fashion given a time budget, 
we record the $\bar{t}_{\mathrm{dis}}$ in Table~\ref{table:quantitative} as the initial planning time.
Each robot now considers robot-robot collision only in the replanning horizon with reduced congestion. As a result, the trajectory optimization computation time $\bar{t}_{\mathrm{traj}}$ reduces significantly compared to the centralized baseline.
Specifically, for the entire trajectory planning computation, i.e., the sum of $\bar{t}_{\mathrm{dis}}$ and $\bar{t}_{\mathrm{traj}}$, in ``Circle74", the proposed algorithm runs  about $13.4$ 
times faster than the centralized method, $64.8$ 
times faster in ``Circle142", $20.7$ 
times faster in ``Horizontal48", and $18.0$ 
times faster in Swap48. Both the average planning time $\bar{t}_{\mathrm{dis}}$, $\bar{t}_{\mathrm{traj}}$ and the averaged maximum planning time $\bar{t}^{\mathrm{max}}_{\mathrm{dis}}$ and $\bar{t}_{\mathrm{traj}}^{\mathrm{max}}$ are within the real-time regime for all instances. 
As expected, while performing in real-time, the algorithm yields suboptimal trajectories according to the average trajectory makespans $\bar{T}$, mainly due to the MCF detouring robots to reduce congestion. This suboptimality is  inevitable but can be controlled by the suboptimality bound $w_{\mathrm{mcf}}$ in MCF routing.

Compared with the decentralized algorithms, our method demonstrates significant advantages in task success rate, leading to collision-avoiding and deadlock-free planning. We report these results in Sec.~\ref{sec:compare_to_centralized_and_decentralized}.

In conclusion, our proposed algorithm has the benefit of real-time replanning capability comparable to a decentralized planner yet achieves a high task success rate comparable to a centralized planner for large-scale swarm trajectory planning tasks.

\subsection{Effectiveness of Partition in Low-level Planning}
\begin{figure}[tb]
    \centering
    \includegraphics[width=0.3\textwidth]{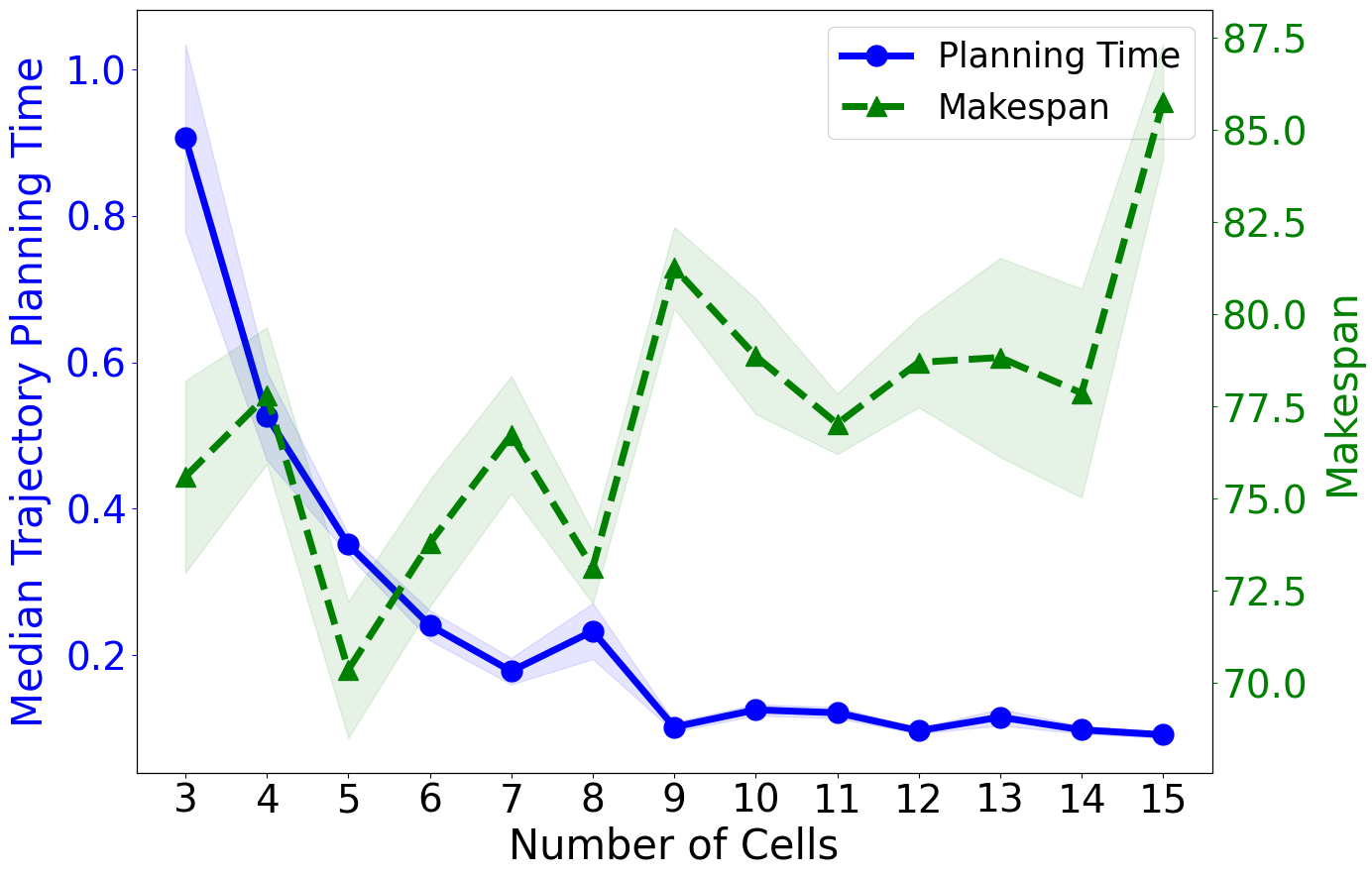}
    \caption{95\% Confidence intervals of median trajectory planning time and makespan of our algorithm with an increasing number of cells. The statistics are averaged over $10$ trials.}
    \label{fig:qualitative_gp}
\end{figure}

\begin{figure*}[t]
    \begin{subfigure}{0.192\textwidth}
       \centering \includegraphics[height=.99\textwidth]{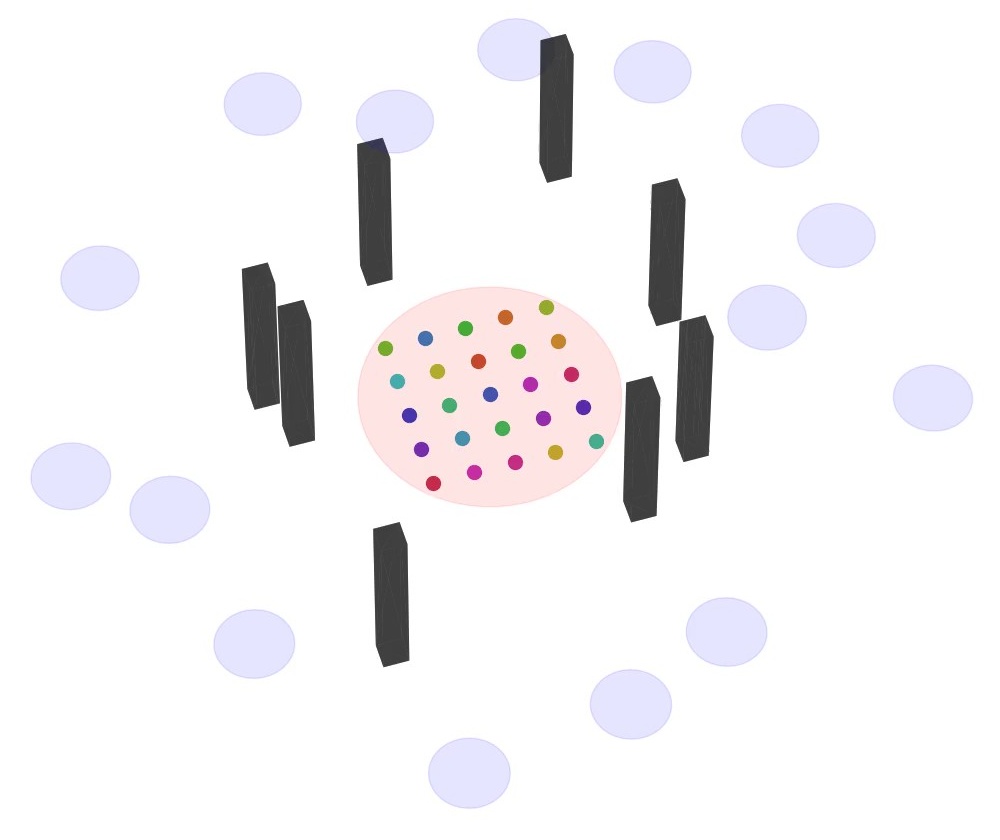}
       \caption{Execution time = 0\unit{s}}
    \end{subfigure}
    \begin{subfigure}{0.192\textwidth}
        \centering \includegraphics[height=.99\textwidth]{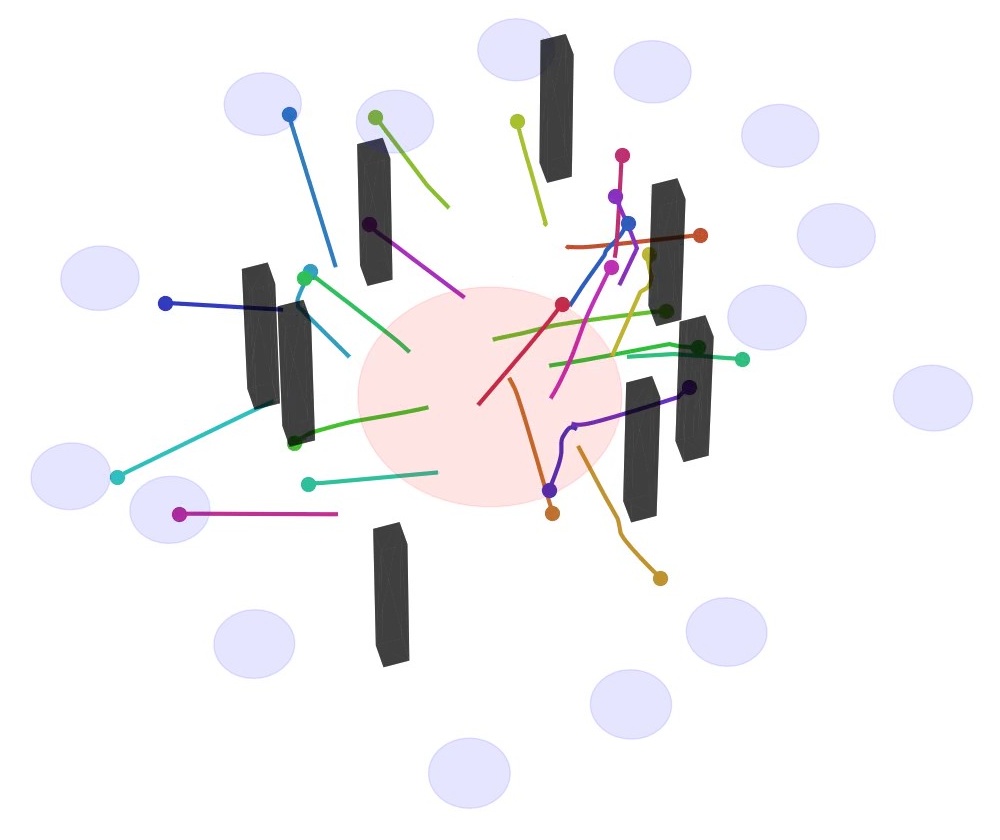}
        \caption{Execution time = 12\unit{s}}
    \end{subfigure}
    \begin{subfigure}{0.192\textwidth}
        \centering \includegraphics[height=.99\textwidth]{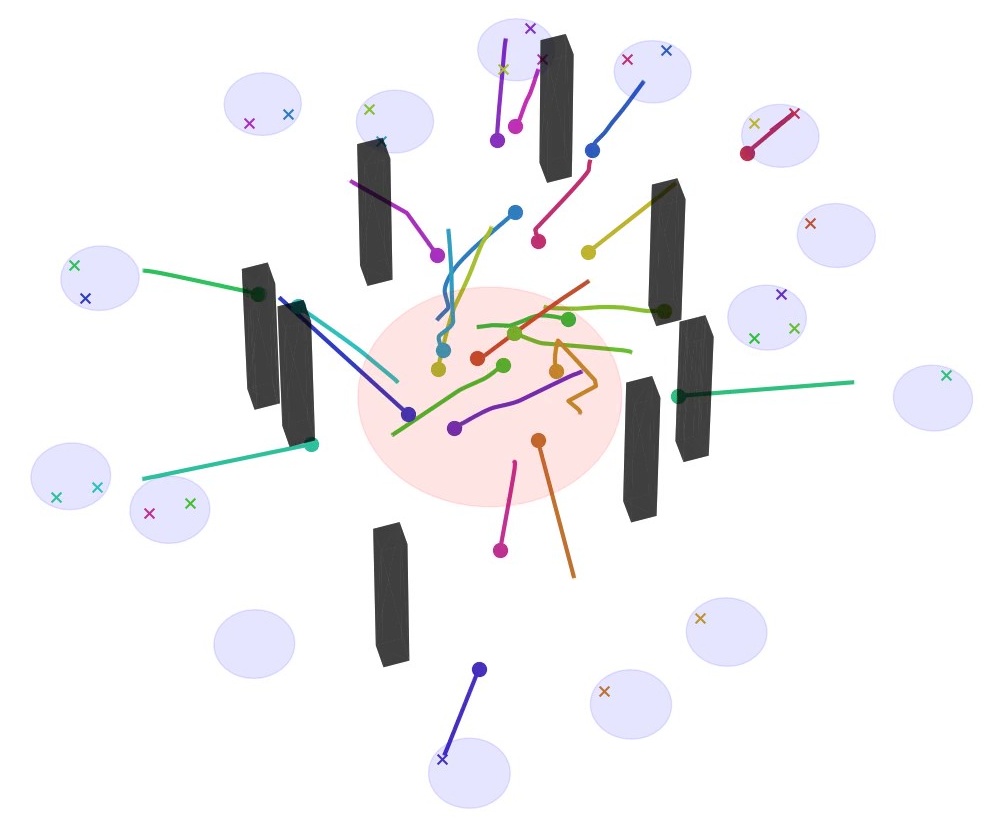}
        \caption{Execution time = 32\unit{s}}
    \end{subfigure}
    \begin{subfigure}{0.192\textwidth}
        \centering \includegraphics[height=.99\textwidth]{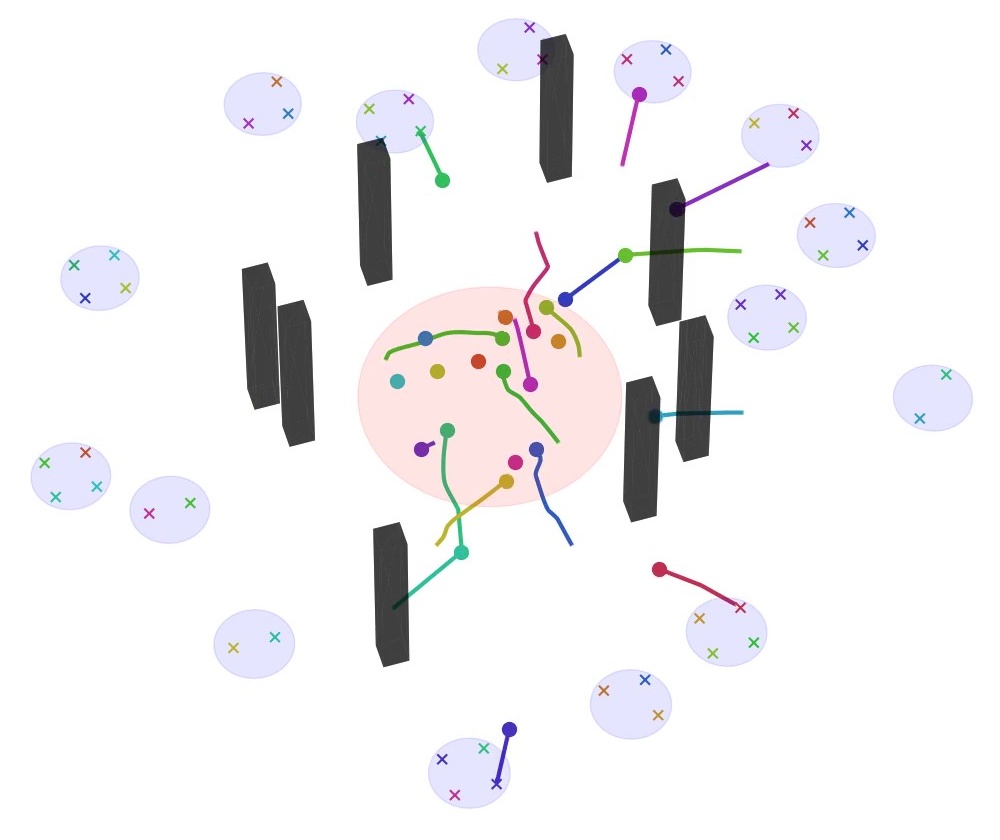}
        \caption{Execution time = 83\unit{s}}
    \end{subfigure}
    \begin{subfigure}{0.192\textwidth}
       \centering \includegraphics[height=.99\textwidth]{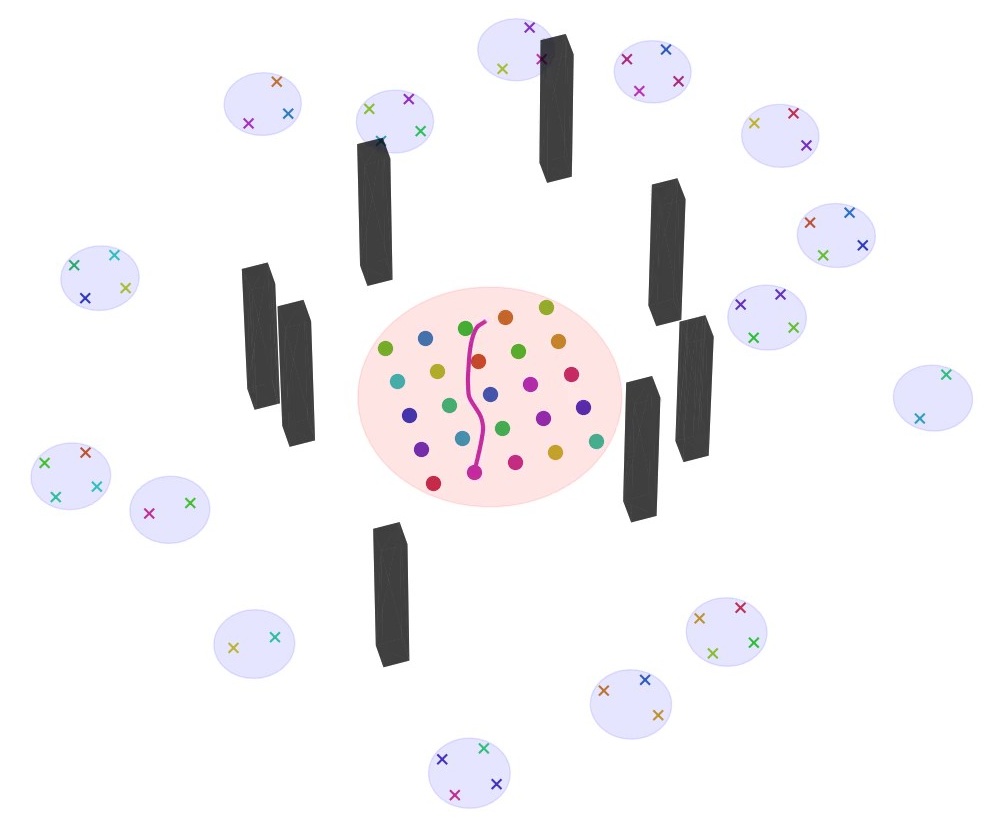}
       \caption{Execution time = 106\unit{s}}
    \end{subfigure}
    \caption{Lifelong experiments with 25 robots. The red circle is the delivery hub, where robots return after each delivery. The blue circles are the access points where packages are dropped. Each robot executes three package deliveries and returns to the hub.}
    \label{fig:lifelong}    
\end{figure*}

\begin{figure*}[!h]
    \centering
    \subfloat[BVC]{\includegraphics[height=0.28\textwidth]{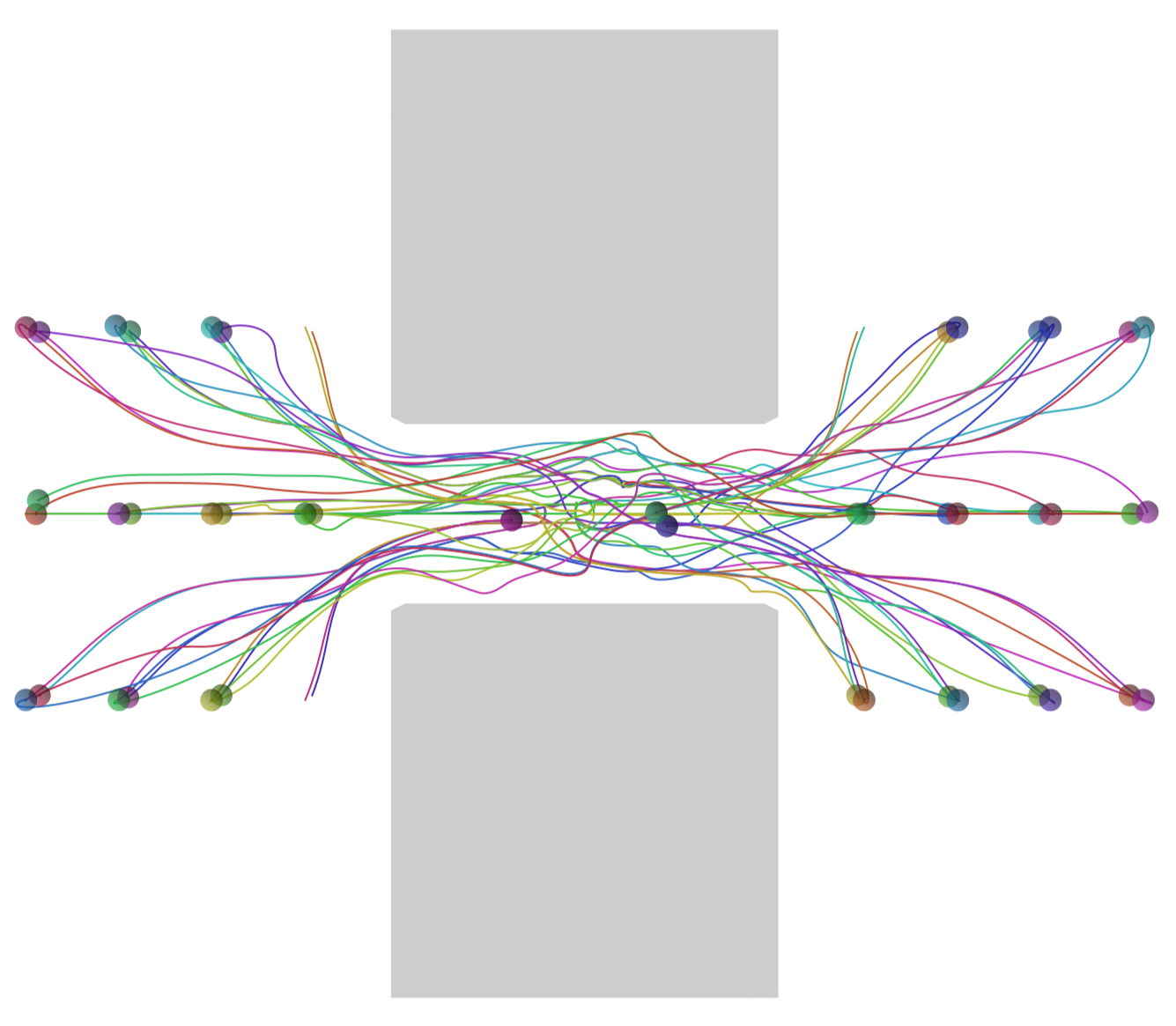}}
    \subfloat[DMPC]{\includegraphics[height=0.28\textwidth]{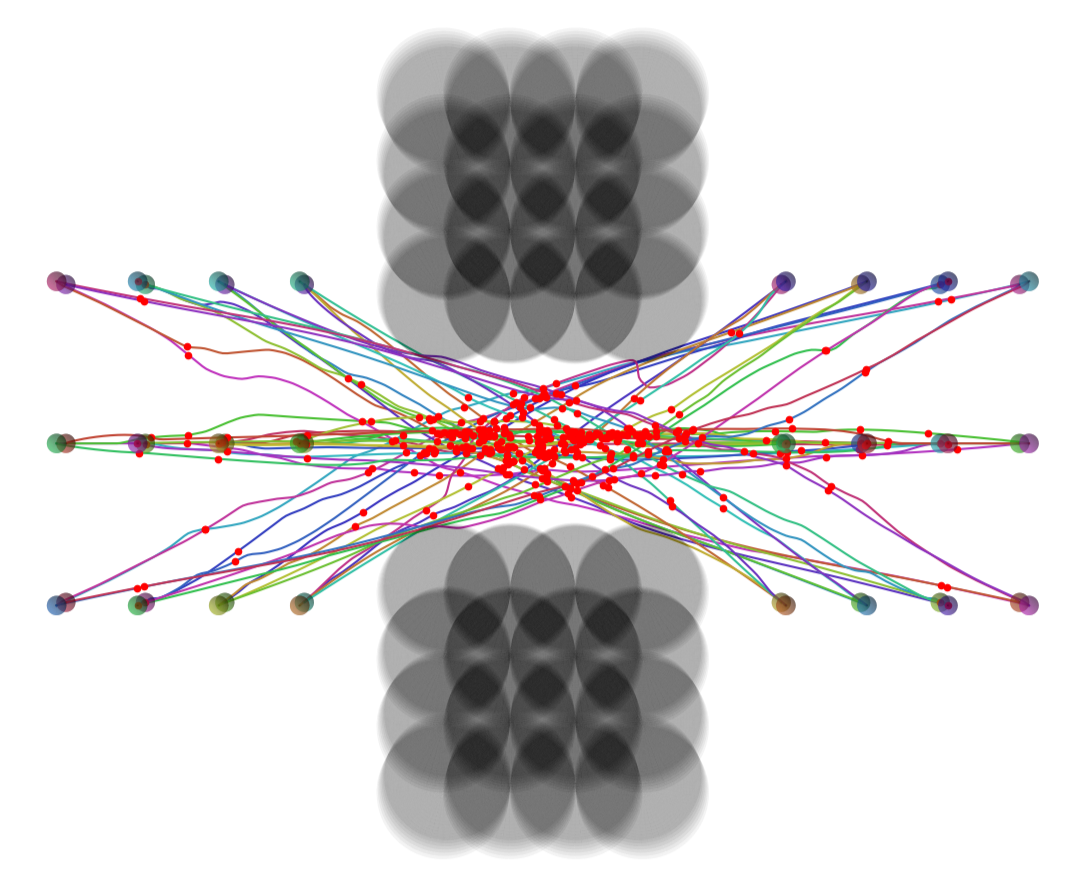}} 
    \subfloat[Ours]{\includegraphics[height=0.28\textwidth]{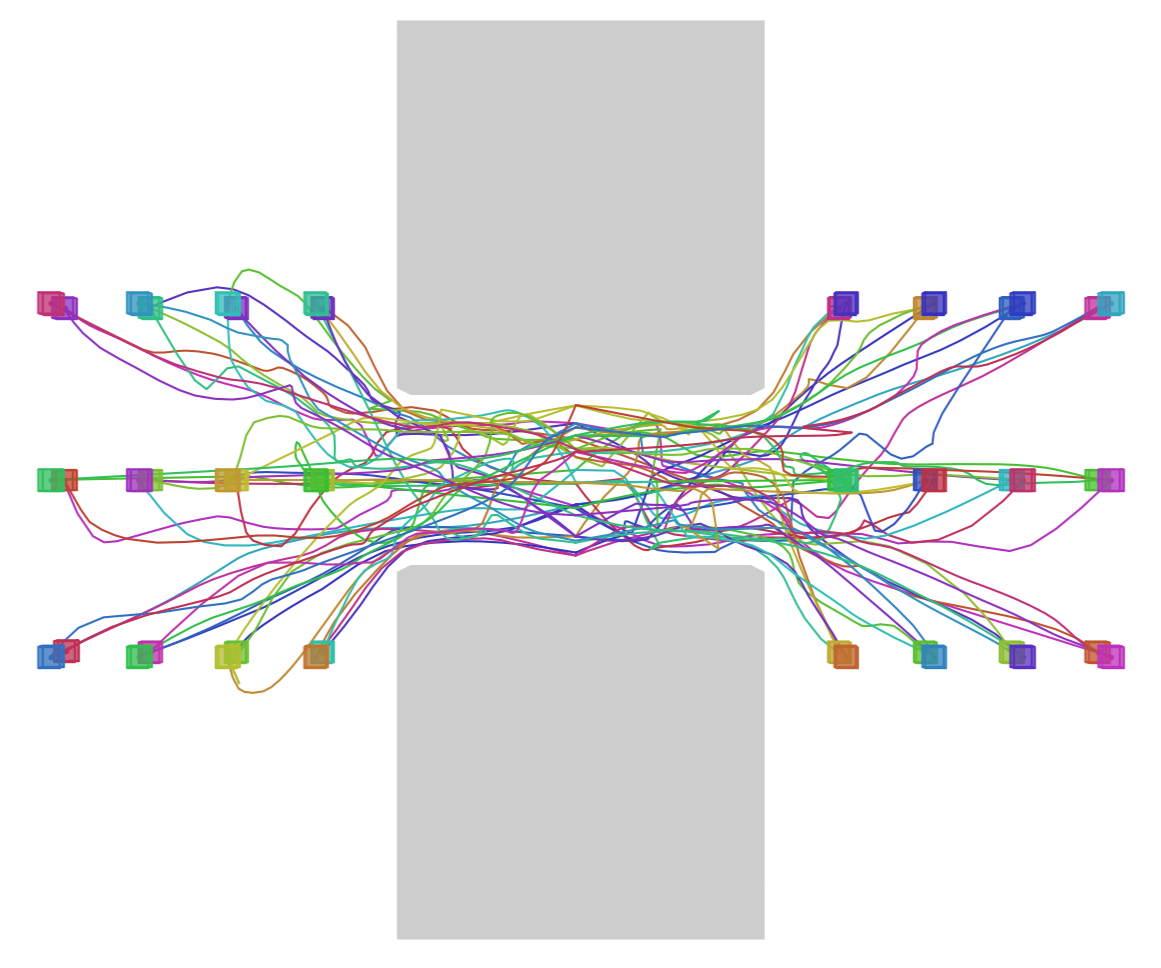}}
    \caption{Typical trajectories for ``Swap48" narrow corridor instance. BVC leads to deadlocks in the corridor (a), DMPC suffers from severe collisions, labeled as red dots (b). Our algorithm completes the task without collision or deadlock. (c)}
    \label{fig:task_completion}
\end{figure*}

We first investigate the influence of the number of cells $Q$. We experiment with a circle of $70$ robots uniformly initialized on a circle with antipodal goals. Fig.~\ref{fig:qualitative_gp} shows  the median trajectory planning time, i.e., the sum of $\bar{t}_{\mathrm{dis}}$ and $\bar{t}_{\mathrm{traj}}$, and trajectory makespan against cell number $Q$. As $Q$ increases,  planning time decreases significantly 
while makespan increases nonlinearly as we increase $Q$. 
As the number of cells increases, the high-level planner becomes more effective in detouring the robots to reduce congestion. However, increasing $Q$ could introduce a shorter inter-cell routing, leading to the nonlinear increase in makespan. Additionally, the partition breaks the global optimality of the MAPF planner, leading to a degenerating solution quality. In summary, by dividing the workspace and parallellizing  computation, our hierarchical approach significantly improves  computational performance with an increase in the makespan.

\subsection{Effectiveness of High-level Planning}

Our MCF-based high-level planner effectively reduces congestion in all cells. We compare the cell robot counts between our MCF-based algorithm and an egocentric greedy approach. The greedy planner outputs a single-robot shortest inter-cell routing, which may lead to cell congestion. Additionally, we compare the MCF-based approach with the centralized baseline by imposing the partition onto the workspace. 

In Table.~\ref{table:quantitative}, we report the average high-level computation time $\bar{t}_{\mathrm{high}}$ and the averaged maximum number of robots in a cell throughout the execution $\bar{N}_{\mathrm{max}}$. 
The complexity of a MAPF instance scales poorly with the number of robots. Thus, $\bar{N}_{\mathrm{max}}$ is an insightful indicator of the computational hardness of a MAPF instance. The computation time for greedy planning is instant, while the centralized MCF-based methods have additional overhead that increases with the number of robots and cells. For all instances in this paper, MCF-based methods provide real-time solutions. 

The MCF-based methods reduce  congestion compared to the greedy and the centralized baseline methods, according to $\bar{N}_{\mathrm{max}}$. 
The MCF-based approach can further reduce the congestion by relaxing the suboptimality bound, i.e., increase $w_{mcf}$. For a fair comparison of low-level planning time, both greedy and MCF-based utilize the same number of threads in computation. Compared to the greedy approach, the proposed MCF-based inter-cell routing brings cell planning to real time while maintaining solution quality, i.e.,  makespan. Note that, for instance ``Circle142", the greedy approach timeouts during discrete planning after $10$ seconds due to congestion in a cell.  
\subsection{Life-long Experiment}
We demonstrate the algorithm's life-long replanning capability in a simulated city-scale drone delivery instance. All UAVs are initialized in a delivery hub colored as a red circle in Fig.~\ref{fig:lifelong}, and asked to deliver multiple packages to access points marked in blue circles. Each UAV delivers three packages and returns to the hub after each successful delivery to recharge and load new packages. Our algorithm successfully completes the lifelong task described above without collision and replans the robots' trajectory in real time. 

\begin{figure*}[!h]
    \centering
    \subfloat[Success rate]{\includegraphics[height=0.24\textwidth]{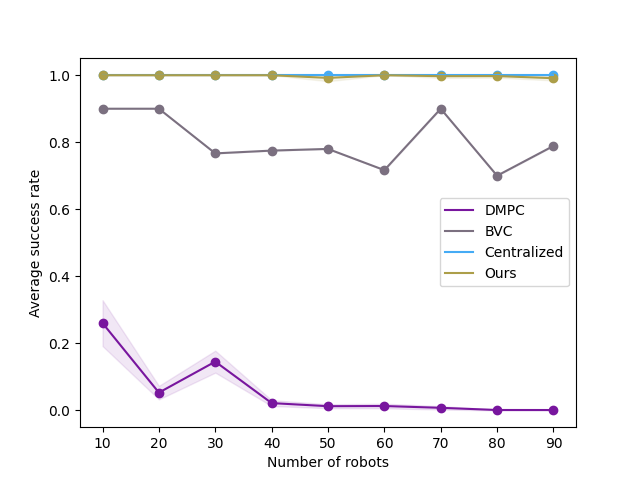}}
    \subfloat[Trajectory Planning Time]{\includegraphics[height=0.24\textwidth]{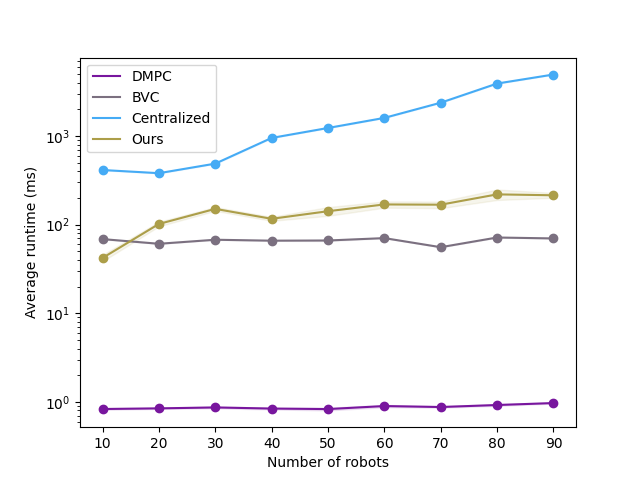}} 
    \subfloat[Makespan]{\includegraphics[height=0.24\textwidth]{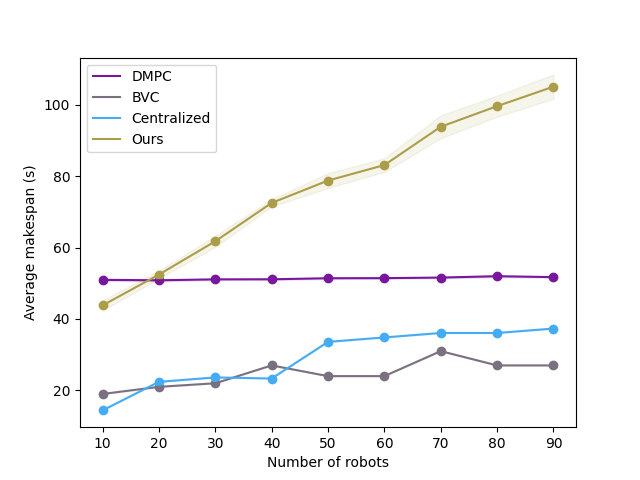}} 
    \caption{Benchmarking with centralized and decentralized approaches. We compare success rate (a), planning computation time (b), and makespan (c).}
    \label{fig:benchmark}
\end{figure*}

\subsection{Comparison with Decentralized and Centralized methods}
\label{sec:compare_to_centralized_and_decentralized}
We compare our hierarchical approach with the SOTA  trajectory planning algorithms, including the decentralized methods  DMPC~\cite{luis2020online} and BVC~\cite{zhou2017fast} and a centralized trajectory planner~\cite{honig2018trajectory}. 
By leveraging MAPF planners in each cell, our algorithm results in deadlock-free trajectories while avoiding collision with neighbors and obstacles. Despite that, in rare cases, we lose collision avoidance guarantee due to optimization infeasibility, our algorithm largely increases the task success rate compared to decentralized algorithms and significantly reduces the computation time compared to a centralized approach. One major drawback of decentralized methods, such as BVC or DMPC, is their inability to complete the task in the presence of narrow corridors with many robots. 

As a qualitative experiment, we swap positions for $48$ robots in a similar ``Swap48" instance to demonstrate the advantage of our algorithm compared to fully decentralized approaches. We use the timestep of $0.1\unit{s}$ for DMPC and $0.2\unit{s}$ for BVC, and adapt the same $1.6\unit{s}$ planning horizon. Our BVC implementation requires a ball collision shape, thus we use a bounding box from $\left[-0.2\unit{m}, -0.2\unit{m}, -0.2\unit{m} \right]^{\top}$ to $\left[0.2\unit{m}, 0.2\unit{m}, 0.2\unit{m} \right]^{\top}$ for collision shape in our algorithm and a ball of radius $0.2\unit{m}$ in both DMPC and BVC. All other parameters for our algorithm are described in Sec.~\ref{sec:simulation_experiments}. All algorithms are implemented in C++. We construct two $24$-robot swarms arranged in a rectangular grid configuration, with $[2\unit{m}, 4\unit{m}, 4\unit{m}]$ spacing in the $x$, $y$, and $z$ axes. The swarms are tasked with swapping positions through a narrow corridor of $8\unit{m}$ in length and $4\unit{m}$ in width. Fig.~\ref{fig:task_completion} illustrates the typical trajectories achieved by BVC, DMPC, and our proposed algorithm. BVC leads to deadlocks in the narrow corridor, as depicted in Fig.~\ref{fig:task_completion}(a). Default DMPC optimizing control inputs with soft collision avoidance constraints results in severe collisions. For each collision, we highlight the centers of mass for collided robots as red dots in Fig.~\ref{fig:task_completion}(b). In comparison, our hierarchical approach results in  trajectories without collisions or deadlocks, as shown in Fig.~\ref{fig:task_completion}(c).

Quantitatively, we compare the success rate, planning time, and solution makespan for our trajectory planning algorithm against decentralized and centralized algorithms with increased numbers of robots. In a workspace of $30$\unit{m}$\times 30$\unit{m}$\times 8$\unit{m} with $20$ randomly generated column obstacles, the number of robots ranging from $10$ to $90$ at increments of $10$ are positioned in a circle with  antipodal goals. We employ $15$ cells in our hierarchical approach. We define a robot as having \emph{successful execution} if it arrives at its goal position without collision. We compute the success rate of a trial as the percentage of robots with successful execution. In Fig.~\ref{fig:benchmark}(a), our hierarchical approach achieves a much higher success rate than the decentralized algorithms BVC and DMPC. Similar to the centralized approach, our algorithm maintains a high task success rate even with increased numbers of robots. In Fig.~\ref{fig:benchmark}(b), our hierarchical approach has a similar computational performance to decentralized methods. It achieves real-time trajectory replanning around $147.2$\unit{ms} across different numbers of robots. Our method demonstrates its significant advantages in planning time compared to the centralized method when the robot number increases. 

In summary, our algorithm achieves computational performance similar to that of decentralized methods while maintaining a high task success rate similar to that of the centralized method. As a trade-off, our algorithm increases the trajectory makespans with increased numbers of robots, shown in Fig.~\ref{fig:benchmark}(c). Despite this increase, we can relax the limit on the number of robots entering one cell to reduce detours, leading to a reduced makespan.

\subsection{Physical Robots}

\begin{figure}[h]
    \centering
    \begin{minipage}{0.24\textwidth}
    \centering
    \includegraphics[width=0.99\textwidth]{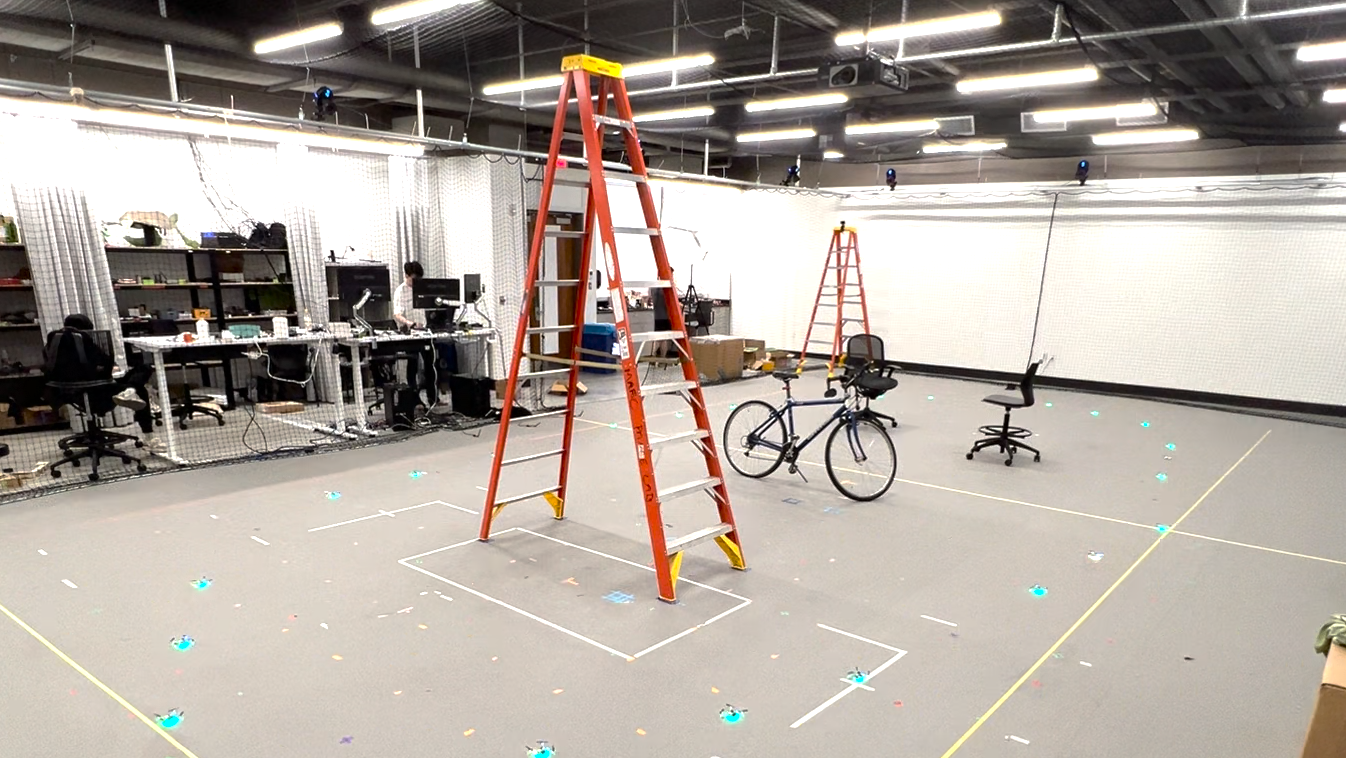}
    \end{minipage}
    \begin{minipage}{0.24\textwidth}
    \centering
    \includegraphics[width=0.99\textwidth]{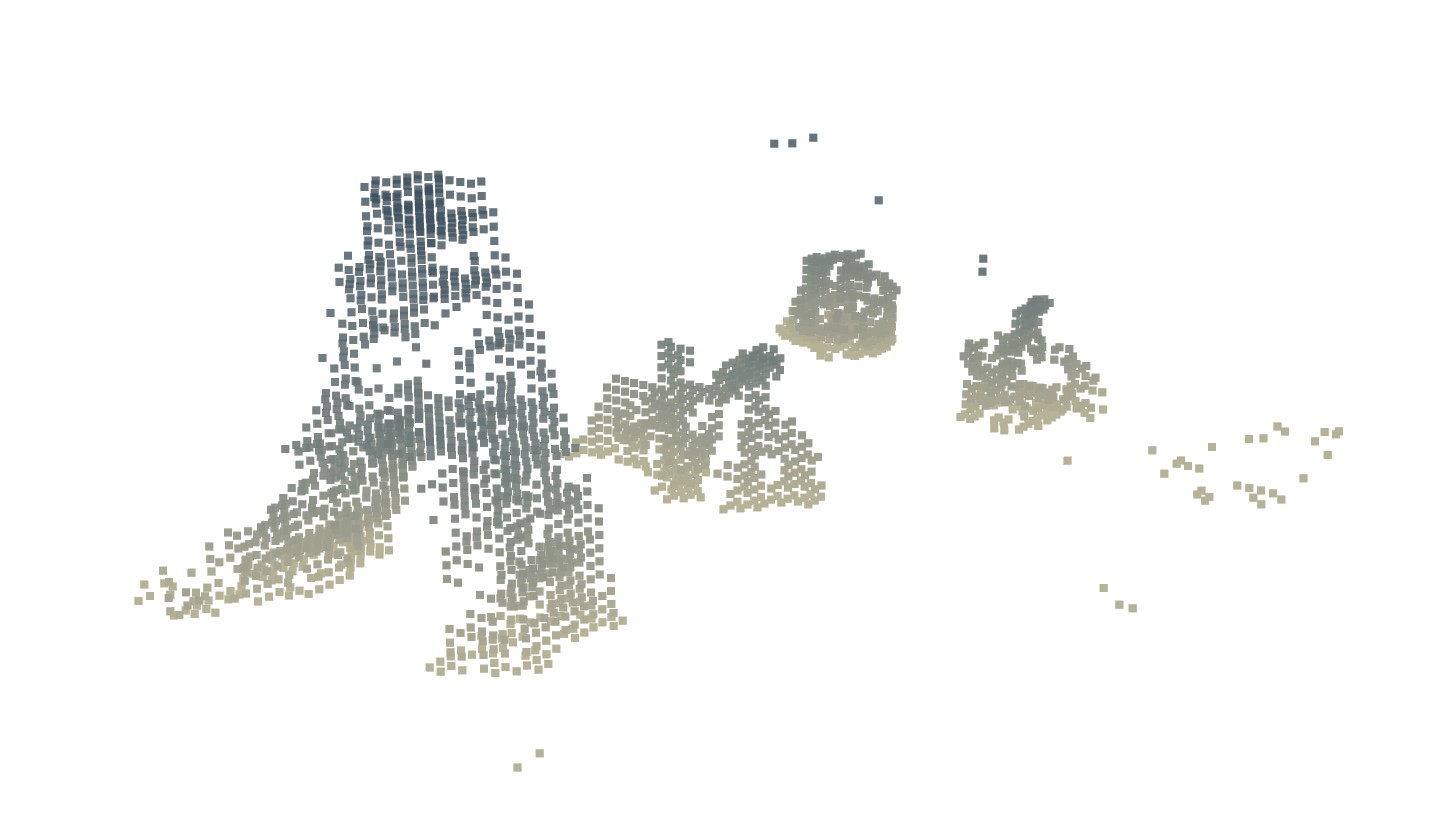}
    \end{minipage}
    \caption{The physical environment with a ladder, a bicycle, two chairs (left), and its Octomap representation (right).}
    \label{fig:actlab_environment}
\end{figure}

We demonstrate our algorithm with a representative $24$ physical Crazyfile nano-quadrotor experiment in a cluttered environment. $24$ Crazyflies are initially uniformly located on an ellipse with antipodal goals. Figure~\ref{fig:actlab_environment} shows the experiment environment with arbitrary obstacles, e.g., a ladder, a bicycle, and two chairs. We use a \textsc{Vicon} motion capture system for localization and CrazySwarm~\cite{preiss2017crazyswarm} to broadcast the control inputs to Crazyfile nano-quadrotors. We place markers on an Intel Realsense D455 depth camera and track its 6-DoF state using \textsc{Vicon} and the \textsc{Octomap} library to create an occupancy map of the $12.55$\unit{m}$\times 7.63$\unit{m}$\times 2.8$\unit{m} lab space. It's worth mentioning that due to the hardware limitations of Crazyflie quadrotors, the trajectories are generated on a centralized computer and broadcast to different drones. However, in principle, the trajectory generation can be done in a distributed manner. 

Figure~\ref{fig:demo} shows the executed trajectories under long-exposure photography. The experiment shows that the proposed algorithm distributes the robots effectively through the workspace (Experiment video link: \url{https://youtu.be/lDi9BYxMf7s}.) 

%% file: sections/conclusion_and_future_work.tex


We have introduced a hierarchical trajectory planning algorithm for large-scale swarm coordination tasks. Our hierarchical approach combines the computational advantage of a decentralized planner, that is, real-time replanning and task performance of a centralized planner, such as deadlock-free and maintaining a high task success rate even for large-scale swarm operations. Our hierarchical framework achieves real-time planning by dividing the workspace into disjoint cells; within each, an anytime MAPF planner computes collision-free paths in parallel. Our high-level planner detours robots and regulates the congestion while guaranteeing routing quality. Despite that the detour yields a suboptimal routing solution, it significantly reduces the task difficulties for both discrete path planning and trajectory optimization by reducing the congestion. Additionally, 
our algorithm considers  robot embodiment, and we run experiments with the downwash-aware collision model of a quadrotor. 
The real-time replanning with deadlock-free and high task success rate operation suits on-demand large-scale applications such as drone delivery. We demonstrate the life-long capability of our algorithm in a city-scale drone delivery example. 

While the MCF-based high-level planner operates in real-time in our experiments with up to $142$ robots, the limits of its real-time operation are not well defined and depend on the number and density of robots and cells in the space.
Future work will explore distributed MCF~\cite{awerbuch2007greedy, awerbuch2012distributed} to increase the scalability of the system. Furthermore, we assume perfect sensing and communication in this work, which can be compromised in realistic systems and operations. In future work, we aim to address large-scale trajectory planning when encountering uncertainty in the localization or unreliable communication between the swarm.  

